\documentclass{article}

\usepackage[utf8]{inputenc} % allow utf-8 input
\usepackage[T1]{fontenc}    % use 8-bit T1 fonts
\usepackage{hyperref}       % hyperlinks
\usepackage{url}            % simple URL typesetting
\usepackage{booktabs}       % professional-quality tables
\usepackage{amsfonts}       % blackboard math symbols
\usepackage{nicefrac}       % compact symbols for 1/2, etc.
\usepackage{microtype}      % microtypography
\usepackage{xcolor}         % colors

\usepackage{epsf}
\usepackage{fancyhdr}
\usepackage{graphics}
\usepackage{graphicx}
\usepackage{psfrag}
\usepackage{microtype}
\usepackage{subcaption}
\usepackage{cancel}
\usepackage{algorithmicx}
\usepackage{color,xcolor}
 \usepackage{algpseudocode}
\usepackage{algorithm}
\algrenewcommand\algorithmicrequire{\textbf{Input:}}
\usepackage{color}

\usepackage{amsthm}
\usepackage{amsfonts}
\usepackage{amsmath}
\usepackage{amssymb,bbm}
\usepackage{comment}
\newcommand*\R[0]{\mathbb{R}}

\newcommand*\D[0]{\mathcal{D}}

\newcommand*\B[0]{\mathcal{B}}

\renewcommand*{\P}{\mathbb{P}}
\DeclareMathOperator*{\argmax}{arg\,max}
\DeclareMathOperator*{\argmin}{arg\,min}

\newcommand{\eps}{\epsilon}

\newcommand{\cN}{\ensuremath{\mathcal{N}}}

\newcommand{\matsnorm}[2]{\left\| #1 \right\|_{{#2}}}

\newcommand{\vecnorm}[1]{\| #1\|}
\newcommand{\enorm}[1]{\vecnorm{#1}} % euclidean norm

\newcommand{\opnorm}[1]{\ensuremath{\matsnorm{#1}{\tiny{\mbox{op}}}}}

\newcommand{\inprod}[2]{\ensuremath{\langle #1 , \, #2 \rangle}}

\newtheoremstyle{named}{}{}{\itshape}{}{\bfseries}{.}{.5em}{\thmnote{#3's }#1}
\theoremstyle{named}

\theoremstyle{plain}

\newlength{\widebarargwidth}
\newlength{\widebarargheight}
\newlength{\widebarargdepth}

\makeatletter
\long\def\@makecaption#1#2{
        \vskip 0.8ex
        \setbox\@tempboxa\hbox{\small {\bf #1:} #2}
        \parindent 1.5em  %% How can we use the global value of this???
        \dimen0=\hsize
        \advance\dimen0 by -3em
        \ifdim \wd\@tempboxa >\dimen0
                \hbox to \hsize{
                        \parindent 0em
                        \hfil
                        \parbox{\dimen0}{\def\baselinestretch{0.96}\small
                                {\bf #1.} #2
                                }
                        \hfil}
        \else \hbox to \hsize{\hfil \box\@tempboxa \hfil}
        \fi
        }
\makeatother

\long\def\comment#1{}
\definecolor{battleshipgrey}{rgb}{0.52, 0.52, 0.51}
\definecolor{darkgray}{rgb}{0.66, 0.66, 0.66}
\definecolor{darkgreen}{rgb}{0.0, 0.2, 0.13}
\definecolor{darkspringgreen}{rgb}{0.09, 0.45, 0.27}
\definecolor{dukeblue}{rgb}{0.0, 0.0, 0.61}
\definecolor{olivedrab7}{rgb}{0.24, 0.2, 0.12}
\definecolor{darkblue}{rgb}{0.0, 0.0, 0.55}
\definecolor{darkscarlet}{rgb}{0.34, 0.01, 0.1}
\definecolor{candyapplered}{rgb}{1.0, 0.03, 0.0}
\definecolor{ao(english)}{rgb}{0.0, 0.5, 0.0}
\definecolor{applegreen}{rgb}{0.55, 0.71, 0.0}

\newcommand{\E}{\mathbb E}

\newcommand{\simiid}{\overset{\mathrm{i.i.d.}}{\sim}}

\newtheorem{example}{Example}

\newtheorem{lemma}{Lemma}
\newtheorem{claim}{Claim}

\newtheorem{corollary}{Corollary}
\newtheorem{definition}{Definition}
\newtheorem{assumption}{Assumption}

\usepackage{mathrsfs}

\usepackage{amsmath,amsfonts,amsthm,amssymb}
\usepackage{graphicx, color, setspace}

\usepackage{comment}

\usepackage{float}
\usepackage{indentfirst}

\newtheorem{theorem}{Theorem}

\long\def\comment#1{}

\long\def\comment#1{}

\newcommand{\IdMat}{\ensuremath{I}}

\newcommand{\thetahat}{\hat\theta_{\mathrm{PO}}}

\usepackage{scalerel}

\newcommand{\Sphere}[1]{\ensuremath{\mathcal{S}^{#1}}}

\newcommand{\M}{\mathcal{M}}
\newcommand{\I}{\mathcal{I}}

\newcommand{\Fit}{\widehat{\texttt{Map}}}

\newcommand{\PR}{\mathrm{PR}}
\newcommand{\thetaPO}{\theta_{\mathrm{PO}}}

\newcommand{\Dexp}{\tilde{\mathcal{D}}}
\newcommand{\LinM}{\mathcal{M}}
\newcommand{\TV}{\mathrm{TV}}

\newcommand{\MapFam}{\boldsymbol{\D}}
\newcommand{\MFun}{{r}}

\newcommand{\ScoreFun}{\phi}
\newcommand{\couple}{\Pi}
\newcommand{\N}{{\mathcal{N}}}

\newcommand{\unif}{{\mathrm{Unif}}}

\newcommand{\Direc}{{u}}
\newcommand{\Direcb}{{v}}

\newcommand{\staterr}{\texttt{StatErr}_n}
\newcommand{\misspecerr}{\texttt{MisspecErr}}

\newcommand{\subgtheta}{\nu_\theta}

\newcommand{\Bern}{\operatorname{Bern}}

\usepackage[numbers]{natbib}

\usepackage{url}% for url's in bib

\usepackage{cleveref}
\usepackage{hyperref}

\usepackage{nicefrac}
\usepackage{comment}

\usepackage{chngpage}

 \usepackage{tabularx}%

\usepackage{enumitem}
\usepackage{booktabs}
\usepackage{caption}

\usepackage{bm,bbm}
\usepackage{mathtools}
\usepackage{appendix}
\usepackage{titletoc}

\theoremstyle{remark}
\newtheorem{remark}{Remark}

\title{Plug-in Performative Optimization}

\author{%
  Licong Lin
  \thanks{
  Department of Statistics, 
  University of California, Berkeley. Email:
  \texttt{liconglin@berkeley.edu}.} 
  \qquad
  Tijana Zrnic \thanks{
 Department of EECS,
  University of California, Berkeley. Email:
  \texttt{tijana.zrnic@berkeley.edu}.}}

\begin{document}

\maketitle

\begin{abstract}
When predictions are performative, the choice of which predictor to deploy influences the distribution of future observations. The overarching goal in learning under performativity is to find a predictor that has low \emph{performative risk}, that is, good performance on its induced distribution. One family of solutions for optimizing the performative risk, including bandits and other derivative-free methods, is agnostic to any structure in the performative feedback, leading to exceedingly slow convergence rates. A complementary family of solutions makes use of explicit \emph{models} for the feedback, such as best-response models in strategic classification, enabling faster rates. However, these rates critically rely on the feedback model being correct. In this work we study a general protocol for making use of possibly misspecified models in performative prediction, called \emph{plug-in performative optimization}. We show this solution can be far superior to model-agnostic strategies, as long as the misspecification is not too extreme. Our results support the hypothesis that models, even if misspecified, can indeed help with learning in performative settings.
\end{abstract}

\section{Introduction}

Predictions have the power to influence the patterns they aim to predict. For example, stock price predictions inform trading decisions and hence prices; traffic predictions influence routing decisions and thus traffic outcomes; recommendations shape users' consumption and thus preferences.

This pervasive phenomenon has been formalized in a framework called \emph{performative prediction}~\citep{perdomo2020performative}. A central feature that distinguishes the framework from traditional supervised learning is the concept of a \emph{distribution map} $\D(\cdot)$. This object, aimed to capture the feedback from predictions to future observations, is a mapping from predictors $f_\theta$ to their induced data distributions $\D(\theta)$. The main goal in performative prediction is thus to deploy a predictor $f_\theta$ that will have good performance after deployment, that is, on its induced distribution $\D(\theta)$. Formally, the goal is to choose predictor parameters $\theta\in\Theta \subseteq \R^{d_\theta}$ so as to minimize the \emph{performative risk}:
\[\PR(\theta) = \E_{z\sim\D(\theta)} [\ell(z;\theta)],\]
where $\ell(z;\theta)$ is the loss incurred by predicting on instance $z$ with model $\theta$. Typically, $z$ is a feature--outcome pair $(x,y)$. We refer to $$\thetaPO = \argmin_{\theta\in\Theta} \PR(\theta)$$ as the \emph{performative optimum}.

The main challenge in optimizing the performative risk lies in the fact that the map $\D(\cdot)$ is not known. We only observe samples from $\D(\theta)$ for models $\theta$ that have been deployed; we do not observe any feedback for the (typically infinitely many) other models. A key discriminating factor between existing solutions for optimizing under performativity is how they cope with this uncertainty.

One group of methods accounts for the feedback without assuming a problem-specific structure for it. This group includes bandit strategies \citep{kleinberg2008multi,jagadeesan2022regret} and derivative-free optimization \citep{Flaxman2004OnlineCO,miller2021outside}. These methods converge to optima at typically slow---without convexity, even exponentially slow---convergence rates. Moreover, their rates rely on regularity conditions that are out of the learner's control, such as convexity of the performative risk \citep{miller2021outside,izzo2021learn, dong2018strategic} or bounded performative effects \citep{jagadeesan2022regret}.

A complementary group of methods---an important starting point for this work---takes feedback into account by positing explicit \emph{models} for it. Such models include best-response models for strategic classification \citep{hardt2016strategic,jagadeesan2021alternative,levanon2021strategic,ghalme2021strategic}, rational-agent models in economics \citep{spence1978job,wooldridge2003reasoning}, and parametric distribution shifts \citep{izzo2021learn, miller2021outside,izzo2022learn}, among others. To argue that methods building on models find optimal solutions, existing analyses assume that the model is \emph{well-specified}. However, models of social behavior are widely acknowledged to be simplistic representations of real-world dynamics.

Yet, despite the unavoidable misspecification of models, they are ubiquitous in practice. Though their simplicity leads to misspecification, it also allows for efficient, interpretable, and practical solutions. Motivated by this observation, in this work we ask: 
\emph{can models for performative feedback be useful, even if misspecified?}

\subsection{Our contribution}

We initiate a study of the benefits of modeling feedback in performative prediction. We show that models---even if misspecified---can indeed help with learning under performativity.

We begin by defining a general protocol for performative optimization with feedback models, which we call \emph{plug-in performative optimization}. The protocol consists of three steps. First, the learner deploys models $\theta_i\sim \Dexp$ and collects data $z_i\sim\D(\theta_i)$, $i\in[n]$. Here, $\Dexp$ is an exploration distribution of the learner's choosing (for example, it can be uniform on $\Theta$ when $\Theta$ is bounded).  The second step is to use the observations $\{(\theta_i,z_i)\}_{i=1}^n$ to fit an estimate of the distribution map. The map is chosen from a parametric family of possible maps $\MapFam_\B = 
\{\D_\beta\}_{\beta\in\B}$, obtained through modeling. The estimation of the map thus reduces to computing an estimate $\hat\beta$. For example, in strategic classification, $\beta$ could be a parameter quantifying the strategic agents' tradeoff between utility and cost. Finally, the third step is to compute the \emph{plug-in performative optimum}:
\[\thetahat = \argmin_{\theta \in\Theta} \PR^{\hat\beta}(\theta) =  \argmin_{\theta \in\Theta} \E_{z\sim\D_{\hat\beta}(\theta)}[\ell(z;\theta)].\]
We prove a general excess-risk bound on $\PR(\thetahat) - \PR(\thetaPO)$, showing that the error decomposes into two terms. The first is a \emph{misspecification error term}, $\misspecerr$, which captures the gap between the true performative risk and the plug-in performative risk $\PR^{\hat\beta}(\theta)$ in the large-sample regime. This term is irreducible and does not vanish as the sample size $n$ grows. The second is a \emph{statistical error term} that captures the imperfection in fitting $\hat\beta$ due to finite samples. For a broad class of problems, our main result can be summarized as follows.

\begin{theorem}[Informal]
\label{thm:informal}
The excess risk of the plug-in performative optimum is bounded by:
$$\PR(\thetahat) - \PR(\thetaPO) \leq c\cdot \textup{\misspecerr} + \tilde O\left(\frac{1}{\sqrt{n}}\right),$$
for some universal constant $c>0$.
\end{theorem}

Therefore, although the misspecification error is irreducible, the statistical error vanishes at a \emph{fast rate}. In contrast, model-agnostic strategies such as bandit algorithms \citep{kleinberg2008multi,jagadeesan2022regret} do not suffer from misspecification but have an exceedingly slow, often exponentially slow, statistical rate. For example, the bandit algorithm of Jagadeesan et al.~\cite{jagadeesan2022regret} has an excess risk of $\tilde O(n^{-\frac{1}{d_\theta+1}})$. This is why feedback models are useful---for a finite $n$, their excess risk can be far smaller than the risk of a model-agnostic strategy due to the rapidly vanishing statistical rate. The statistical rate is fast because it only depends on the parametric estimation rate of $\hat\beta$; it does not depend on the complexity of $\PR$.

One important case of performative prediction is \emph{strategic classification}. We apply our general theory to common best-response models in strategic classification. We also conduct numerical evaluations that confirm our theoretical findings. Overall our results support the use of models in optimization under performative feedback.

\subsection{Related work}

% We give an overview of existing threads most closely related to our work.

\paragraph{Performative prediction.} We build on the growing body of work studying performative prediction~\citep{perdomo2020performative}. Existing work studies different variants of retraining \citep{perdomo2020performative,mendler2020stochastic,drusvyatskiy2022stochastic}, which converge to so-called performatively stable solutions, as well as methods for finding performative optima \citep{miller2021outside,izzo2021learn,jagadeesan2022regret}. The methods in the latter category are largely model-agnostic and as such converge at slow rates. Exceptions include the study of parametric distribution shifts \citep{izzo2021learn,izzo2022learn} and location families \citep{miller2021outside,jagadeesan2022regret}, but those analyses crucially rely on the model being well-specified. We are mainly interested in settings where $\D(\theta)$ is a general black-box. Other work in performative prediction includes the study of time-varying shifts \citep{brown2022performative,izzo2022learn,li2022state,ray2022decision}, multi-agent settings \citep{dean2022multi,li2022multi,narang2022learning,piliouras2022multi}, causality and robustness \citep{maheshwari2022zeroth,mendler2022anticipating,kim2022making}, and it would be valuable to extend our theory on the use of models to those settings.

\paragraph{Strategic classification and economic modeling.} 
Strategic classification
\citep{hardt2016strategic,dong2018strategic,levanon2021strategic, zrnic2021leads}, as well as other problems studying strategic agent behavior, frequently use models of agent behavior in order to compute Stackelberg equilibria, which are direct analogues of performative optima. However, convergence to Stackelberg equilibria assumes correctness of the models, a challenge we circumvent in this work. We use strategic classification as a primary domain of application of our general theory.

\paragraph{Statistics under model misspecification.} Our work is partially inspired by works in statistics studying the benefits and impact of modeling, including under misspecification \citep{white::1980,white1982maximum,buja2019models1,buja2019models2}. At a technical level, our results are related to M-estimation~\citep{van2000asymptotic,geer2000empirical,mou2019diffusion} and  semiparametric statistics~\citep{tsiatis2007semiparametric,kennedy2022semiparametric}.
%where the goal is to find models that lead to minimal estimation error.

\paragraph{Zeroth-order optimization.} Plug-in performative optimization serves as an alternative to black-box baselines for zeroth-order optimization, which have previously been studied in performative prediction. These include bandit algorithms \citep{kleinberg2008multi,jagadeesan2022regret} and zeroth-order convex optimization algorithms \citep{Flaxman2004OnlineCO,miller2021outside}. As mentioned, we show that the use of models can give far smaller excess risk, given the fast convergence rates of parametric learning problems.

\section{Plug-in performative optimization protocol}

We describe the main protocol at the focus of our study and then instantiate it with an example.

We consider the use of parametric models $\MapFam_{\B} := \{\D_{\beta}\}_{\beta\in\B}$ for modeling the true unknown distribution map $\D$, where $\B\subseteq \R^{d_\beta}$.
We denote 
$$\PR^\beta(\theta) = \E_{z\sim\D_\beta(\theta)} [\ell(z;\theta)].$$ 
Since $\MapFam_\B$ is a collection of maps, we call it a \emph{distribution atlas}. We emphasize that it need not hold that $\D \in \MapFam_\B$; the model could be misspecified.

The protocol for plug-in performative optimization proceeds as follows.
First, the learner collects pairs of i.i.d. observations $\{(\theta_i,z_i)\}_{i=1}^n$, where $\theta_i$ is deployed according to some exploration distribution $\Dexp$ and
$z_i\sim\D(\theta_i)$. The exploration distribution should be ``dispersed enough'' to enable capturing varied distributions $\D(\theta_i)$ (e.g., uniform, Gaussian with a full-rank covariance, etc). Then, the learner estimates the distribution map by fitting $\hat\beta$ based on the collected observations: $$\hat\beta = \Fit((\theta_1,z_1),\dots,(\theta_n,z_n)),$$
where $\Fit$ is some model-fitting function.
We will consider different criteria for fitting $\hat\beta$. We let $\beta^*$ denote the large-sample limit of $\hat\beta$, $\beta^* = \lim_{n\rightarrow\infty}\hat\beta$. Finally, the learner finds the \emph{plug-in performative optimum}:
  \begin{align*}\thetahat = \argmin_{\theta\in\Theta} \PR^{\hat\beta}(\theta) = \argmin_{\theta\in\Theta} \E_{z\sim\D_{\hat\beta}(\theta)} [\ell(z;\theta)].
  % \label{eq:thetahatpo_def}
  \end{align*}

We summarize the protocol in Algorithm~\ref{alg:general_procedure}.

Notice that, since  $\D_{\hat\beta}$ is known to the learner, we may solve for $\hat{\theta}_{\mathrm{PO}}$ explicitly in Step 3 of the protocol, without collecting any additional real data. In particular, solving for $\hat{\theta}_{\mathrm{PO}}$ incurs only computational complexity---\emph{not} statistical complexity. A detailed discussion on how to execute Step 3 empirically can be found in Appendix~\ref{app:optimization}.

\begin{algorithm}[t]
\caption{Plug-in performative optimization}\label{alg:general_procedure}
\begin{algorithmic}[1]
\Require{distribution atlas $\MapFam_\B$, exploration strategy $\Dexp$, loss $\ell(z;\theta)$, map-fitting algorithm $\Fit$.}
\State Deploy $\theta_i\sim\Dexp$, observe $z_i \sim \D(\theta_i)$, $i\in[n]$.
\State Fit distribution map: $\hat\beta = \Fit((\theta_1,z_1),\dots,(\theta_n,z_n))$, where $\hat\beta\in\B$. 
\State Compute plug-in performative optimum: $\thetahat = \argmin_{\theta\in\Theta} \E_{z\sim\D_{\hat\beta}(\theta)} [\ell(z;\theta)]$.
\end{algorithmic}
\end{algorithm}

A canonical choice of \textup{$\Fit$} that we will focus on is \emph{empirical risk minimization}: $$\hat \beta = \argmin_{\beta\in\B} \frac 1 n \sum_{i=1}^n \MFun(\theta_i,z_i;\beta),$$
where 
$\MFun$ is a loss function. Throughout we will use $\tilde\theta$ and $\tilde z$ to denote draws $\tilde\theta\sim\Dexp,\tilde z\sim\D(\tilde\theta)$. Then, $\beta^* = \argmin_{\beta\in\B} \E[\MFun(\tilde\theta,\tilde z;\beta)]$.  For example, one can choose $\MFun(\theta,z;\beta) = -\log p_\beta(z;\theta)$ to be the log-likelihood, where $p_\beta(\cdot;\theta)$ is the density under $\D_\beta(\theta)$, in which case $\hat\beta$ is the maximum-likelihood estimator.
Under this choice, $$\beta^* = \argmax_{\beta\in\B} \E [\log p_\beta (\tilde z;\tilde \theta)] = \argmin_{\beta\in\B} \mathrm{KL}(\bar\D,\bar\D_\beta).$$
Here, $\bar\D$ is the distribution of $\tilde z$, that is, the distribution map $\D(\theta)$ averaged over $\theta\sim\Dexp$. We similarly define $\bar\D_\beta$. Therefore, $\beta^*$ is the KL projection of the true data-generating process onto the considered distribution atlas.

\begin{example}[Biased coin flip]\label{exm:biased_coin} To build intuition for the introduced concepts, we consider an illustrative example. Suppose we want to predict the outcome of a biased coin flip, where the bias arises due to performative effects. The outcome is generated as $z\sim\D(\theta) = \Bern(0.5+\mu\theta+\eta\theta^2)$, where $\mu\in(0,0.5),\eta\in(0,0.5-\mu)$. The parameter $\theta\in[0,1]$ aims to predict the outcome while minimizing the squared loss, $\ell(z;\theta) = (z-\theta)^2$.
Suppose that we know that $\theta$ introduces a bias to the coin flip, but we do not know how strongly or in what way. We thus choose a simple model for the bias,  $\D_\beta(\theta) = \Bern(0.5+\beta\theta)$, and fit $\beta$ in a data-driven way. To do so, we deploy $\theta_i\simiid\unif[0,1]$ and observe $z_i\sim\D(\theta_i)$, for $i\in[n]$. One natural way to fit the distribution map is to solve
$$\hat\beta=\argmin_{\beta}\frac{1}{n}\sum_{i=1}^n(z_i-0.5 - \beta\theta_i)^2.$$
Finally, we compute the plug-in performative optimum as 
$$\thetahat = \argmin_\theta \E_{z\sim\D_{\hat\beta}(\theta)} [(z-\theta)^2] = \frac{1-\hat\beta}{2-4\hat\beta}.$$
It is not hard to show that the population limit of $\hat\beta$ is equal to $\beta^* = \mu + 0.75\eta$. Therefore, if the feedback model is well-specified, meaning $\eta = 0$, then $\beta^*$ indeed recovers the true distribution map, and $\thetahat$ converges to the true performative optimum.
\end{example}

\section{Excess risk}

We study the excess risk of plug-in performative optimization. The key takeaway of this section is that the excess risk depends on two sources of error: one is the \emph{misspecification error} due to the fact that, often, $\D\not\in\MapFam_\B$; the other is the \emph{statistical error} due to the gap between $\hat\beta$ and $\beta^*$.

Formally, define
\begin{align*}\misspecerr &= \sup_{\theta\in\Theta} |\PR^{\beta^*}(\theta) - \PR(\theta)|,\\
 \staterr &= \sup_{\theta\in\Theta} |\PR^{\beta^*}(\theta) - \PR^{\hat\beta}(\theta)|.
 \end{align*}
We note that the statistical error depends on the sample size $n$, while the misspecification error is irreducible even in the large-sample limit. In later sections we will show that the statistical error vanishes at a \emph{fast rate}, namely $\tilde O\left(\frac{1}{\sqrt{n}}\right)$, for a broad class of problems. In Theorem \ref{thm:general_risk_bound} we state a general bound on the excess risk in terms of these two sources of error.

\begin{theorem}
\label{thm:general_risk_bound}
The excess risk of the plug-in performative optimum is bounded by:\[\PR(\thetahat) - \PR(\thetaPO) \leq 2\left(\textup{\misspecerr} + \textup{\texttt{StatErr}}_n\right).\]
\end{theorem}

Theorem \ref{thm:general_risk_bound} illuminates the benefits of feedback models: if the model is a reasonable approximation, the misspecification error is not too large; at the same time, due to the parametric specification of the distribution atlas, the statistical error vanishes quickly. Therefore, we conclude that \emph{even misspecified} models can lead to lower excess risk than entirely model-agnostic strategies such as bandit algorithms. 

\begin{remark}
It should be noted that there may be numerical inaccuracy in solving for $\thetahat$ in Step 3 of Algorithm~\ref{alg:general_procedure}. However, the bound of Theorem~\ref{thm:general_risk_bound} degrades smoothly: if a $\delta$-suboptimal solution is attained, then the excess risk increases by at most~$\delta$. As mentioned before, $\delta$ is not dependent on $n$; it only depends on the amount of computation.
\end{remark}

% The error decomposition of Theorem \ref{thm:general_risk_bound} is reminiscent of the classical bias--variance tradeoff in statistics. When the atlas $\MapFam$ is very expressive, the misspecification error is low but the statistical error is high---similar to having low bias and high variance. Similarly, when $\MapFam$ is coarse, the misspecification error is high but the statistical error is low---similar to having large bias and low variance.

In the rest of this section we give fine-grained bounds on the misspecification error and the statistical error under appropriate regularity assumptions, providing intuition via examples along the way. 
The most natural way to bound the misspecification error is in terms of a distributional distance between the true distribution map $\D(\theta)$ and the modeled distribution map $\D_{\beta^*}(\theta)$. We define the misspecification of a distribution atlas.

\begin{definition}[Misspecification]
We say that a distribution atlas is $\eta$-misspecified in distance $\textup{\texttt{dist}}$ if, for all $\theta\in\Theta$, it holds that $\textup{\texttt{dist}}(\D_{\beta^*}(\theta),\D(\theta))\leq \eta$.
\end{definition}

We will measure misspecification in either total-variation distance or Wasserstein (i.e. earth mover's) distance. Depending on the problem setting, one of the two distances will yield a smaller misspecification parameter and thus a tighter rate according to Theorem \ref{thm:general_risk_bound}.

We will also require that the atlas is ``smooth'' in the analyzed distance.

\begin{definition}[Smoothness]
\label{def:lipschitz_model}
We say that a distribution atlas is $\epsilon$-smooth in distance $\textup{\texttt{dist}}$ if, for all $\beta,\beta'\in\B$ and $\theta\in\Theta$, it holds that $\textup{\texttt{dist}}(\D_{\beta}(\theta),\D_{\beta'}(\theta))\leq \epsilon\|\beta-\beta'\|$.
\end{definition}

Unless stated otherwise, $\|\cdot\|$ denotes the $\ell_2$-norm. In some examples the parameter of the atlas will be a matrix, in which case the norm will be the operator norm $\opnorm{\cdot}$. It is important to note that, while the misspecification parameter is a property of the true distribution map, the smoothness parameter is entirely in the learner's control, as it is solely a property of the chosen distribution atlas.

In what follows, Section \ref{sec:tv_misspec} and Section \ref{sec:wass_misspec} focus on bounding the misspecification error. Section~\ref{sec:est_rate} focuses on bounding the statistical error with an explicit rate.

\subsection{Total-variation misspecification}
\label{sec:tv_misspec}
First we consider misspecification in total-variation (TV) distance. Building on Theorem~\ref{thm:general_risk_bound}, we obtain the following excess-risk bound as a function of TV misspecification.

\begin{corollary}
\label{cor:tv_main}
Suppose the distribution atlas is $\eta_{\TV}$-misspecified and $\epsilon_{\TV}$-smooth in TV distance. Moreover, suppose that $|\ell(z;\theta)|\leq B_\ell$ and $\|\hat \beta - \beta^*\|\leq C_n$. Then, the excess risk of the plug-in performative optimum is bounded~by:
\[\PR(\thetahat) - \PR(\thetaPO) \leq 4B_\ell \cdot\eta_{\TV} + 4B_\ell \cdot \epsilon_{\TV}\cdot C_n.\]
\end{corollary}

Corollary \ref{cor:tv_main} shows that plug-in performative optimization is efficient as long as the distribution atlas is smooth, not too misspecified, and the rate of estimation $C_n$ is fast. In Section~\ref{sec:est_rate} we will prove convergence rates $C_n$ when $\hat\beta$ is a sufficiently regular empirical risk minimizer.

We now build intuition for the relevant terms in Corollary~\ref{cor:tv_main} through examples. First we give a couple of examples of distribution atlases and bound their smoothness parameter~$\epsilon_{\mathrm{TV}}$.

\begin{example}[Mixture model]
Suppose that we have $k$ candidate distribution maps $\{\D^{(i)}(\theta)\}_{i=1}^k$. We would like to find a combination of these maps that approximates the true map $\D(\theta)$ as closely as possible. To do so, we can define
$\D_\beta(\theta) = \sum_{i=1}^k \beta_i \D^{(i)}(\theta)$, where $\beta\in[0,1]^k$ defines the mixture weights. This model is smooth in TV distance: $\TV(\D_{\beta}(\theta), \D_{\beta'}(\theta)) \leq \frac{1}{2}\|\beta - \beta'\|_1 \leq \frac{\sqrt{k}}{2} \|\beta - \beta'\|_2$.
\end{example}

\begin{example}[Self-fulfilling prophecy]
\label{ex:prophecy}
Suppose that we want to model outcomes that follow a ``self-fulfilling prophecy,'' meaning that predicting a certain outcome makes it more likely for the outcome to occur. Assume we have historical data of feature--label pairs before the model was deployed. Denote the resulting empirical distribution by $\D_0 = \D_{0}^X \times \D_0^{Y|X}$. We assume the features are nonperformative; only the labels exhibit performative effects. Then, we can model the label distribution as $\D_\beta^{Y|X}(\theta) = (1-\beta)\D_0^{Y|X} + \beta \delta_{f_\theta(X)}$, where $\delta_{f_\theta(X)}$ denotes a point mass at the predicted label. Here, $\beta\in[0,1]$ tunes the strength of performativity: $\beta = 0$ implies no performativity, while $\beta = 1$ a perfect self-fulfilling prophecy. This atlas has TV-smoothness equal to $\epsilon_{\TV}=1$.
% : $\TV(\D_{\beta}(\theta), \D_{\beta'}(\theta)) \leq |\beta - \beta'|$.
\end{example}
Next, we describe a general type of misspecification that implies a bound on $\eta_{\mathrm{TV}}$.

\begin{example}[``Typically'' well-specified model]
Suppose that the data distribution consists of observations about strategic individuals. Suppose that a $(1-p)$-fraction of the population is ``rational'' and acts in a predictable fashion. The remaining $p$-fraction acts arbitrarily. Then, if we model the predictable behavior appropriately, meaning $\D_{\beta^*}(\theta)$ follows the distribution produced by the rational agents, the misspecification parameter $\eta_{\mathrm{TV}}$ is at most $p$. More generally, if we have $\D(\theta) = (1-p)\D_{\beta^*}(\theta) + p\tilde \D(\theta)$, where $\tilde \D(\theta)$ is an arbitrary component, then $\eta_{\mathrm{TV}} \leq p$.
\end{example}

\subsection{Wasserstein misspecification}
\label{sec:wass_misspec}

We next consider misspecification in Wasserstein (i.e. earth mover's) distance. Building on Theorem~\ref{thm:general_risk_bound}, we bound the excess-risk via Wasserstein misspecification.

\begin{corollary}
\label{cor:wass_main}
Suppose that the distribution atlas is $\eta_{W}$-misspecified and $\epsilon_{W}$-smooth in Wasserstein distance. Moreover, suppose that the loss $\ell(z;\theta)$ is $L_z$-Lipschitz in $z$, and that $\|\hat\beta - \beta^*\|\leq C_n$. Then, the excess risk of the plug-in performative optimum is bounded by:
\[\PR(\thetahat) - \PR(\thetaPO) \leq 2L_z \cdot\eta_{W} + 2L_z \cdot\epsilon_W \cdot C_n.\]
\end{corollary}

As in Corollary \ref{cor:tv_main}, we see that the excess risk of the plug-in performative optimum is small as long as the distribution atlas is smooth, not overly misspecified, and the rate $C_n$ is sufficiently fast. 

Below we give an example of a natural distribution atlas and characterize its Wasserstein smoothness.

\begin{example}[Performative outcomes]
\label{ex:perf_outcomes}
As in Example \ref{ex:prophecy}, suppose that we have data of feature--outcome pairs before any model deployment, and suppose that only the outcomes are performative while the features are static.
Let $\D_0$ denote the historical data distribution. We assume that a predictor $\theta$ affects the outcomes only through its predictions $f_\theta(x)$. One simple way to model such feedback is via an additive effect on the outcomes. Formally, we define $(x,y)\sim\D_\beta(\theta) \Leftrightarrow (x_0,y_0)\sim \D_0, x = x_0, y=y_0 + \beta \cdot f_\theta(x)$. As in Example \ref{ex:prophecy}, $\beta\in\R$ controls the strength of performativity. This atlas is $\epsilon_W$-smooth in Wasserstein distance for $\epsilon_W = \sup_{\theta} \E_{x\sim\D_0^X} [|f_\theta(x)|]$.
\end{example}

One way that misspecification can arise is due to \emph{omitted-variable bias}. We illustrate this in the following example and explicitly bound the misspecification parameter $\eta_W$. 

\begin{example}[Omitted-variable bias]
Suppose that only a subset of the coordinates $\I\subseteq [d]$ of $\theta$ induce performative effects. This can happen in linear or logistic regression, where the coordinates of $\theta$ measure feature importance, but only a subset of the features are manipulable. Specifically, assume the data follows a \emph{location family} model: $z\sim \D(\theta)\Leftrightarrow z = z_0 + \tilde\M \theta_{\I}$, where $\tilde \M \in~\R^{d_z\times |\I|}$ is a true parameter of the shift and $z_0$ is a zero-mean draw from a base distribution $\D_0$. Suppose the model omits one performative coordinate by mistake: $z\sim\D_\M(\theta)\Leftrightarrow z = z_0 + \M \theta_{\I'}$, where $\I' = \I\setminus \{i_{\mathrm{miss}}\}$ and $\M\in\R^{d_z\times |\I'|}$. If $\widehat\M$ is fit via least-squares, $\widehat \M = \argmin_\M \frac{1}{n}\sum_{i=1}^n\|z_i - \M \theta_{i,\I'}\|^2$, and $\Dexp$ is a product distribution, then the population-level counterpart of $\widehat \M$ is equal to $\M^* = \tilde \M_{\I'}$, which denotes the restriction of $\tilde\M$ to the columns indexed by $\I'$. Putting everything together, we can conclude that the misspecification due to the omitted coordinate is $\eta_W = \sup_{\theta} \mathcal W(\D(\theta),\D_{\M^*}(\theta)) \leq B\|\tilde \M_{i_{\mathrm{miss}}}\|$, where we assume the $i_{\mathrm{miss}}$-coordinate of $|\theta|$ is at most $B$.

% |\theta_{i_{\mathrm{miss}}}|\leq B, \forall\theta$.
% % We will return to location families in Section \ref{sec:location_fam}.
\end{example}

\subsection{Bounding the estimation error}
\label{sec:est_rate}

We saw that the statistical error is driven by the estimation rate of $\beta^*$. We show that for a broad class of problems the rate is $\tilde O(n^{-\frac{1}{2}})$. We focus on map-fitting via \emph{empirical risk minimization}~(ERM):
\begin{equation}
\label{eq:ERM}
\hat\beta = \argmin_{\beta\in\B} \frac 1 n \sum_{i=1}^n \MFun(\theta_i,z_i;\beta),
\end{equation}
where $\B\subset \R^{d_\beta}$ is bounded and convex.
We denote
$\MFun(\beta) =\E[\MFun(\tilde \theta,\tilde z;\beta)]$, and note that $\beta^* = \argmin_{\beta\in\B} r(\beta)$. 
To establish the error bound, we rely on the following assumptions.

\begin{assumption}
\label{ass:regularity}
ERM problem~\eqref{eq:ERM} is regular if:
\begin{enumerate}[label=(\alph*)]
\itemsep0em 
% \item
% \label{assn:m_est_finite_regu}
% The set $\B$ is bounded and convex.
% \item
% \label{assn:m_est_smooth}
% $\MFun(\theta,z;\beta)$ is twice continuously differentiable in $\beta$.
\item
\label{assn:m_est_finite_cvx}
% There exists $\mu>0$ such that
% $\nabla \MFun(\beta)^\top(\beta-\beta^*) \geq\mu \|\beta-\beta^*\|^2$, for all $\beta\in\B$.
$\MFun(\beta)$ is convex and $\mu$-strongly convex at $\beta^*$, and the Hessian $\nabla^2\MFun(\beta)$ is $\sigma_r$-Lipschitz;
\item
\label{assn:m_est_finite_bnd}
$\forall \beta\in\B$, the gradient
$\nabla \MFun(\tilde \theta,\tilde z;\beta)$ is a subexponential vector with  parameter $B_\MFun>0$;  %,\theta\in\Theta$. 
\item 
\label{assn:m_est_finite_lip}
$\forall u,u'\in \mathcal S^{d_\beta-1}$,
 %and $\theta\in\Theta$,
$\sup_{\beta\in\B} u^\top\nabla^2\MFun(\tilde \theta,\tilde z;\beta)u'$ is subexponential with  parameter $L_\MFun>0$. 
\end{enumerate}
\end{assumption}

We will see that Assumption \ref{ass:regularity} is satisfied in many important settings. Condition (a) is fairly standard; conditions (b,c) seem less standard because we want to make them realistic
for problems with diverging dimension. They immediately hold if $\nabla r(\theta,z;\beta),\nabla^2 r(\theta,z;\beta)$ are bounded, which are standard conditions.
% , and can be easily verified in the location-family application, for example.

The following lemma is the key tool for obtaining our main excess-risk bounds with a \emph{fast} rate, stated in Theorem~\ref{thm:risk_bound_w_est_rate}.

\begin{lemma}
\label{lemma:estimation_lemma}
Assume the map-fitting algorithm is regular (Ass.~\ref{ass:regularity}) and that $\frac{n}{\log n} \geq C (d_\beta + \log(1/\delta))$ for a sufficiently large  $C>0$. Then, for some constant $C'>0$, with probability $1-\delta$ it holds that
$$\|\hat\beta - \beta^*\|\leq C'\sqrt{\log n} \sqrt{\frac{d_\beta + \log(1/\delta)}{n}}.$$
\end{lemma}

% Combined with Corollary \ref{cor:tv_main} and Corollary \ref{cor:wass_main}, Lemma \ref{lemma:estimation_lemma} implies our main excess-risk bounds.

\begin{theorem}
\label{thm:risk_bound_w_est_rate}
Assume the map-fitting algorithm is regular (Ass.~\ref{ass:regularity}) and that $\frac{n}{\log n} \geq C (d_\beta + \log(1/\delta))$ for a sufficiently large  $C>0$.
\begin{itemize}
\item If the distribution atlas is $\eta_{\TV}$-misspecified and $\epsilon_{\TV}$-smooth in total-variation distance, and $|\ell(z;\theta)|\leq B_\ell$, then there exists a $C'>0$ such that, with probability $1-\delta$:
\begin{align*}
\PR(\thetahat) - \PR(\thetaPO) \leq 4B_\ell \eta_{\TV} + C' B_\ell   \epsilon_{\TV}  \sqrt{\log n} \sqrt{\frac{d_\beta + \log(1/\delta)}{n}}.
\end{align*}
\item If the distribution atlas is $\eta_W$-misspecified and $\epsilon_W$-smooth in Wasserstein distance, and $\ell(z;\theta)$ is $L_z$-Lipschitz in $z$, then there exists a $C'>0$ such that, with probability $1-\delta$:
\begin{align*}
\PR(\thetahat) - \PR(\thetaPO) \leq 
2L_z \eta_W +  C' L_z  \epsilon_W  \sqrt{\log n} \sqrt{\frac{d_\beta + \log(1/\delta)}{n}}.
\end{align*}
\end{itemize}
\end{theorem}

The excess risk is thus bounded by the sum of a term due to misspecification and a \emph{fast} statistical rate. To contrast this with a model-agnostic rate, the bandit algorithm of~\cite{jagadeesan2022regret} does not suffer from misspecification but has an \emph{exponentially} slow excess risk, $\tilde O(n^{-1/(d_\theta+1)})$.

The analysis underlying Theorem \ref{thm:risk_bound_w_est_rate} is based on standard ERM analyses under model misspecification, though there are differences. The main one is that our setting requires that we analyze the error in terms of the difference between parameters $\hat\beta-\beta^*$, as opposed to the excess risk $r(\hat\beta)-r(\beta^*)$ in the standard ERM analysis. This difference requires new tools and assumptions akin to those in Assumption~\ref{ass:regularity}.

\section{Applications}
We apply our theory and  plug-in performative optimization to  several problems with performative feedback, building on prevalent models for those problems. 
In each problem, we prove the model's smoothness and fast estimation of $\beta^*$.
% In each problem, we show that the models are smooth and achieve fast estimation of $\beta^*$.

\subsection{Strategic regression}\label{sec:stra_regr}

We begin by considering strategic regression. Here, a population of individuals described by $(x,y)$ strategically responds to a deployed predictor $f_\theta$. For example, the predictor could be $f_\theta(x)=\theta^\top x$.

\paragraph{Distribution atlas.} The strategic responses consist of manipulating features in order to maximize a utility function, which is often equal to the prediction itself. Formally, given an individual with features $x_0$, a commonly studied response model~is
$g_\beta(x_0,\theta) =  \argmax_{x} (u_\theta(x) - \frac{1}{2\beta} \|x-x_0\|^2 ),$
where $u_\theta(x)$ is a concave utility function and the second term captures the cost of feature manipulations. Here, $\beta\in [\beta_{\min},\beta_{\max}]\subseteq \R_+$ trades off utility and cost. The natural distribution atlas capturing the above response model is obtained as follows. Suppose that we have a historical distribution of feature--label pairs $\D_0$. Then, let:
\begin{align}
\label{eq:strat_reg_atlas}
\hspace{-0.2cm}(x,y)\sim \D_\beta(\theta)
\Leftrightarrow (x_0,y)\sim \D_0, x = g_\beta(x_0,\theta).
\end{align}

\begin{claim}
\label{claim:strat_reg_smoothness}
If $\|\nabla u_\theta(x)\|\leq B_u$ and $\nabla u_\theta(x)$ is $L_u$-Lipschitz, then $\{\D_\beta\}_\beta$  has $$\epsilon_W \leq \frac{B_u}{1-\beta_{\max}L_u}.$$
\end{claim}

This bound is attained for $u_\theta(x) = x^\top \theta$. There, $L_u = 0,B_u = \sup_{\theta\in\Theta} \|\theta\|$, so the bound equals $\epsilon_W\leq \max_{\theta\in\Theta} \|\theta\|$. This is tight because $\sup_{x,\theta\in\Theta} \|g_\beta(x,\theta) - g_{\beta'}(x,\theta)\| = |\beta-\beta'|\sup_{\theta\in\Theta} \|\theta\|$.

\paragraph{Map fitting.} We can fit $\hat \beta$ via maximum-likelihood estimation (MLE). Suppose that $\D_{0}^X$ has a density and denote it by $p_{0}^X$. Then, we can let 
$$\hat\beta = \argmin_{\beta\in\B} -\frac 1 n \sum_{i=1}^n \log\left(p_{0}^X(x_i - \beta\nabla u_{\theta_i}(x_i))\right).$$
Given the first-order optimality condition for $g_\beta(x_0,\theta)$, under mild regularity and correct specification of the model, meaning $\D(\theta) = \D_\beta(\theta)$ for some $\beta$, this map-fitting strategy ensures $\eta_W = 0$, as expected. For example, if $\D_{0}^X = \N(0,\sigma^2 \IdMat)$, MLE reduces to $$\hat\beta = \argmin_{\beta\in\B} \frac 1 n \sum_{i=1}^n \|x_i - \beta\nabla u_{\theta_i}(x_i)\|^2.$$
Least-squares makes sense even if the features are not Gaussian; it just coincides with MLE for Gaussians. We show a fast estimation rate of least-squares under mild conditions via Lemma \ref{lemma:estimation_lemma}.

% In particular, the first-order optimality condition for the response model is equal to:
% $$x_0 = g_\beta(x_0,\theta) - \beta \nabla u_\theta(g_\beta(x_0,\theta)).$$

% If the model is well-specified, then $\beta^*$ recovers the true model and is unique under mild assumptions.

% Note that the squared error objective makes sense regardless, only it may not recover the true parameter $\beta_0$ in general.

\begin{claim}
\label{claim:strat_reg_rate}
If $\E [\|\nabla u_{\tilde \theta}(\tilde x)\|^2]>0$, $\tilde x$ and $\nabla u_{\tilde \theta}(\tilde x)$ are subgaussian, and $\frac{n}{\log n}\geq C(1 + \log(1/\delta))$ for a sufficiently large $C>0$, then 
$$|\hat\beta-\beta^*| \leq C' \sqrt{\log n}\sqrt{\frac{ 1 + \log(1/\delta)}{n}}$$ with probability $1-\delta$.
\end{claim}

If, in addition, the loss function $\ell(z;\theta)$ is $L_z$-Lipschitz in $z$, combining Claim~\ref{claim:strat_reg_smoothness}, Claim~\ref{claim:strat_reg_rate}, and 
Corollary~\ref{cor:wass_main} gives an upper bound on the excess risk $\PR(\thetahat)-\PR(\thetaPO)$.

\subsection{Binary strategic classification}

Next, we consider binary strategic classification, in which a population of strategic individuals described by $(x,y)$ takes strategic actions in order to reach a decision boundary. We assume the learner's decision rule is obtained by thresholding a linear model, $f_\theta(x) = \mathbf{1}\{\theta^\top x \geq T\}$, for some $T$. Without loss of generality we assume $\|\theta\|=1$, since the rule is invariant to rescaling $\theta$ and $T$.

\paragraph{Distribution atlas.} A common model assumes that the individuals have a budget $\beta>0$ on how much they can change their features \citep{kleinberg2020classifiers,chen2020learning,zrnic2021leads}. The individuals move to the decision boundary if it is within $\ell_2$ distance $\beta$. Formally, an individual with features $x_0$ responds with
$g_\beta(x_0,\theta) =
x_0 + \theta(T - x_0^\top \theta)$ if   $x_0^\top \theta \in [T - \beta,T)$, and does not move otherwise.
The natural distribution atlas corresponding to the above model is defined as in Eq.~\eqref{eq:strat_reg_atlas},
% \begin{align}
% \label{eq:binary_class_atlas}
% (x,y)\sim\D_\beta(\theta) \Leftrightarrow (x_0,y_0)\sim\D_0, y = y_0, x=g_\beta(x_0,\theta),
% \end{align}
for a given base distribution $\D_0$. We show that this atlas is smooth in total-variation distance.

\begin{claim}
\label{claim:strat_class_smoothness}
If, for all $\theta$, $x_0^\top \theta$ has a density upper bounded by $\phi_u$, then $\{\D_\beta\}_\beta$ has $\epsilon_{\mathrm{TV}} \leq \phi_u$.
\end{claim}

\paragraph{Map fitting.} According to the atlas, all individuals with $x_0^\top\theta \in [T-\beta,T)$ move to the decision boundary, defined by $x^\top \theta = T$. Therefore, one can estimate the individuals' budget by finding $\hat \beta$ such that $\P\{x_0^\top \tilde \theta \in [T- \hat \beta,T]\} = \frac{1}{n} \sum_{i=1}^n \mathbf{1}\{x_i^\top \theta_i \in (T\pm\epsilon)\}$,
for a small $\epsilon>0$. The latter term estimates the mass in a small neighborhood of the boundary. Therefore, $\P\{x_0^\top \tilde \theta \in [T-  \beta^*,T]\} = \P\{\tilde x^\top \tilde \theta \in (T\pm\epsilon)\}$. For simplicity we assume $x_0^\top\tilde \theta$ has a density on $\R$ so that $\hat \beta$ exists. 
Note that $\P\{x_0^\top \tilde \theta \in [T- \beta,T]\}$ is known for all $\beta$ because it is a property of the base distribution, so finding $\hat \beta$ reduces to estimating $\P\{\tilde x^\top \tilde \theta \in (T\pm\epsilon)\}$.

\begin{claim}
\label{claim:strat_class_rate}
If $x_0^\top \tilde \theta$ has a density lower bounded by $\phi_l$, then 
$$|\hat \beta - \beta^*|\leq \frac{1}{\phi_l} \sqrt{\frac{\log(2/\delta)}{2n}}$$
with probability $1-\delta$.
\end{claim}

If the learner's loss is bounded, putting together Claim~\ref{claim:strat_class_smoothness}, Claim~\ref{claim:strat_class_rate}, and 
Corollary~\ref{cor:tv_main} gives an upper bound on the excess risk $\PR(\thetahat)-\PR(\thetaPO)$.

\subsection{Location families}
\label{sec:location_fam}

Lastly, we consider general location families \citep{miller2021outside,jagadeesan2022regret,ray2022decision}, in which the deployment of $\theta$ leads to performativity via a linear shift. This model often appears in strategic classification with linear or logistic regression, and can capture performativity only in certain features~\citep{miller2021outside}.

\paragraph{Distribution atlas.} The location-family model is defined by $z\sim \D_{\LinM}(\theta) \Leftrightarrow z = \LinM\theta+z_0$,
where $\LinM\in\R^{d_z\times d_\theta}$ is a matrix that parameterizes the shift, and $z_0$ is a sample from a zero-mean base distribution $\D_0$. We assume $\sup_{\theta\in \Theta}\|\theta\|\leq B_\theta$. It is not hard to see that the atlas is smooth in $\opnorm{\cdot}$.

\begin{claim}
\label{claim:loc_family_1}
The atlas $\{\D_\LinM\}_{\LinM}$ has $\epsilon_W \leq B_\theta$.
\end{claim}

\paragraph{Map fitting.} We fit the distribution map via least-squares:
$$\widehat\LinM
= \argmin_{\LinM} \frac{1}{n}\sum_{i=1}^n \|z_i-\LinM\theta_i\|^2.$$
Thus, $\LinM^* = \argmin_\LinM \E[\|\tilde z-\LinM\tilde \theta\|^2]$.
We provide control on the estimation error below.

\begin{claim}\label{claim:loc_family_2}
Assume $\Dexp$ is zero-mean and subgaussian with $\kappa_{\min}\IdMat \preceq\E[\tilde \theta\tilde \theta^\top]
 \preceq \kappa_{\max}\IdMat$. Further, for all $u\sim\mathcal S^{d_\theta-1}, v\sim\mathcal S^{d_z-1}$, assume $u^\top \tilde \theta \cdot v^\top \tilde z$ is subexponential with parameter $L_{\theta z}$ and $\|\E[\tilde \theta \tilde z^T]\|_{\mathrm{op}}\leq B$. Then, if $n\geq C(d_\theta+d_z+\log(1/\delta))$ for some sufficiently large $C>0$, there exists $C'>0$ such that with probability $1-\delta$ we have
   $$\|\widehat\LinM-\LinM^*\|_{\mathrm{op}}\leq C'\sqrt{\frac{d_\theta+d_z+\log(1/\delta)}{n}}.$$
\end{claim}

The above assumptions hold under mild regularity  conditions when the model is nearly well-specified. If the loss is $L_z$-Lipschitz in $z$, then using Claim~\ref{claim:loc_family_1}, Claim~\ref{claim:loc_family_2}, and Corollary~\ref{cor:wass_main}, we can bound the excess performative risk $\PR(\thetahat)-\PR(\thetaPO)$.

\newcommand{\incp}{{b}}

\section{Experiments}
\label{sec:experiments}
We confirm the qualitative takeaways of our theory empirically.  
 We compare our approach with two model-agnostic algorithms: derivative-free optimization (DFO)~\cite{Flaxman2004OnlineCO} and greedy SGD~\citep{mendler2020stochastic}. The latter naively retrains while ignoring the feedback; it is a practical heuristic but only converges to stable points.
For the location-family experiment, we also compare our approach with PerfGD~\cite{izzo2021learn}, which approximates the performative gradient via numerical methods. We refer the reader to Appendix \ref{app:experiments} for further details.

% We confirm the qualitative takeaways of our theory empirically. 
% % We consider two settings: location families and strategic regression.
% We compare our method with two model-agnostic strategies: the derivative-free optimization (DFO) method  by~\cite{Flaxman2004OnlineCO}, and a method that simply retrains and ignores feedback, in particular greedy SGD~\citep{mendler2020stochastic}. Additional details can be found in the Appendix.

\subsection{Location family}

We start with the location-family setting. We assume the true map follows a linear model with a quadratic term, $z_i = \incp+\LinM_1\theta_i+s \LinM_2(\theta_i\circ\theta_i)+z_{0,i}$,
where $z_i,\theta_i,\incp \in\R^{d}$,
 $\theta_i\circ\theta_i:= (\theta_{i,1}^2,\ldots,\theta_{i,d}^2)$ represents the quadratic effect, and $z_{0,i}\sim\N(0,\sigma^2\IdMat_d)$. The parameter $s\geq 0$ varies the magnitude of misspecification. We want to minimize the loss $\ell(z;\theta)=\|z-\theta\|^2$, and we use a simple linear model to approximate $\D(\theta)$, i.e., $z\sim\D_\LinM(\theta) \Leftrightarrow z\overset{d}{=} \incp+\LinM\theta+~z_0$.
To fit $\widehat\LinM$, we use the loss $\MFun(\theta,z;\LinM)=\enorm{z-\LinM\theta}^2$.
We vary $d\in\{5,10\}$ and let $\LinM_i=\tilde \LinM_i/\|\tilde \LinM_i\|_{\mathrm{op}},~i\in\{1,2\}$ where $\tilde\LinM_i\in\R^{d\times d}$ have entries generated i.i.d. from $\N(0,1)$. We let $\incp\sim \N(0,I_d)$, $\sigma=0.5$, and $\Theta = \{\theta:\|\theta\|\leq 1\}$.

\begin{figure}[t]
  \centering
     \includegraphics[width=0.3\textwidth]{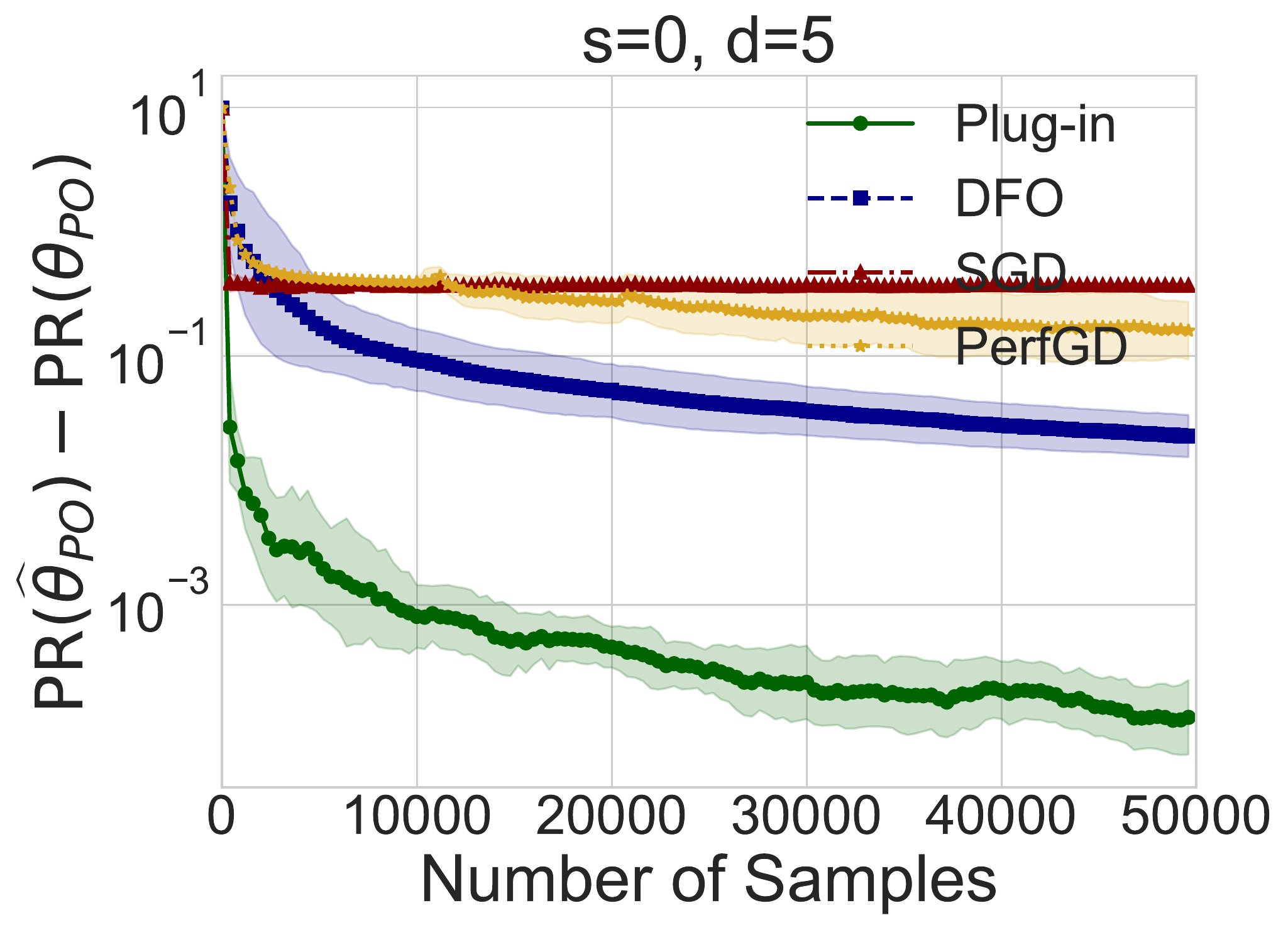}
  \includegraphics[width=0.3\textwidth]{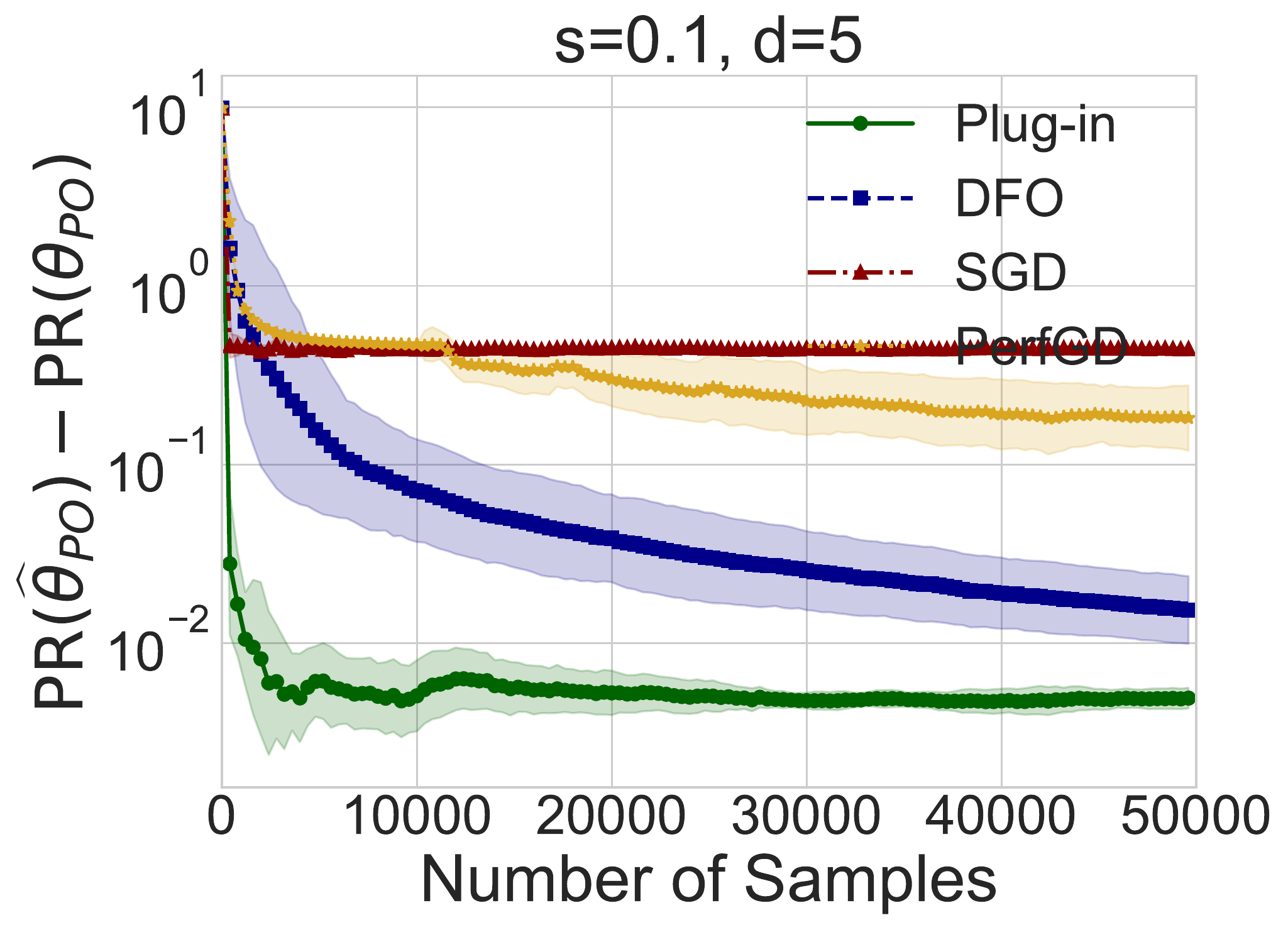}
    \includegraphics[width=0.3\textwidth]{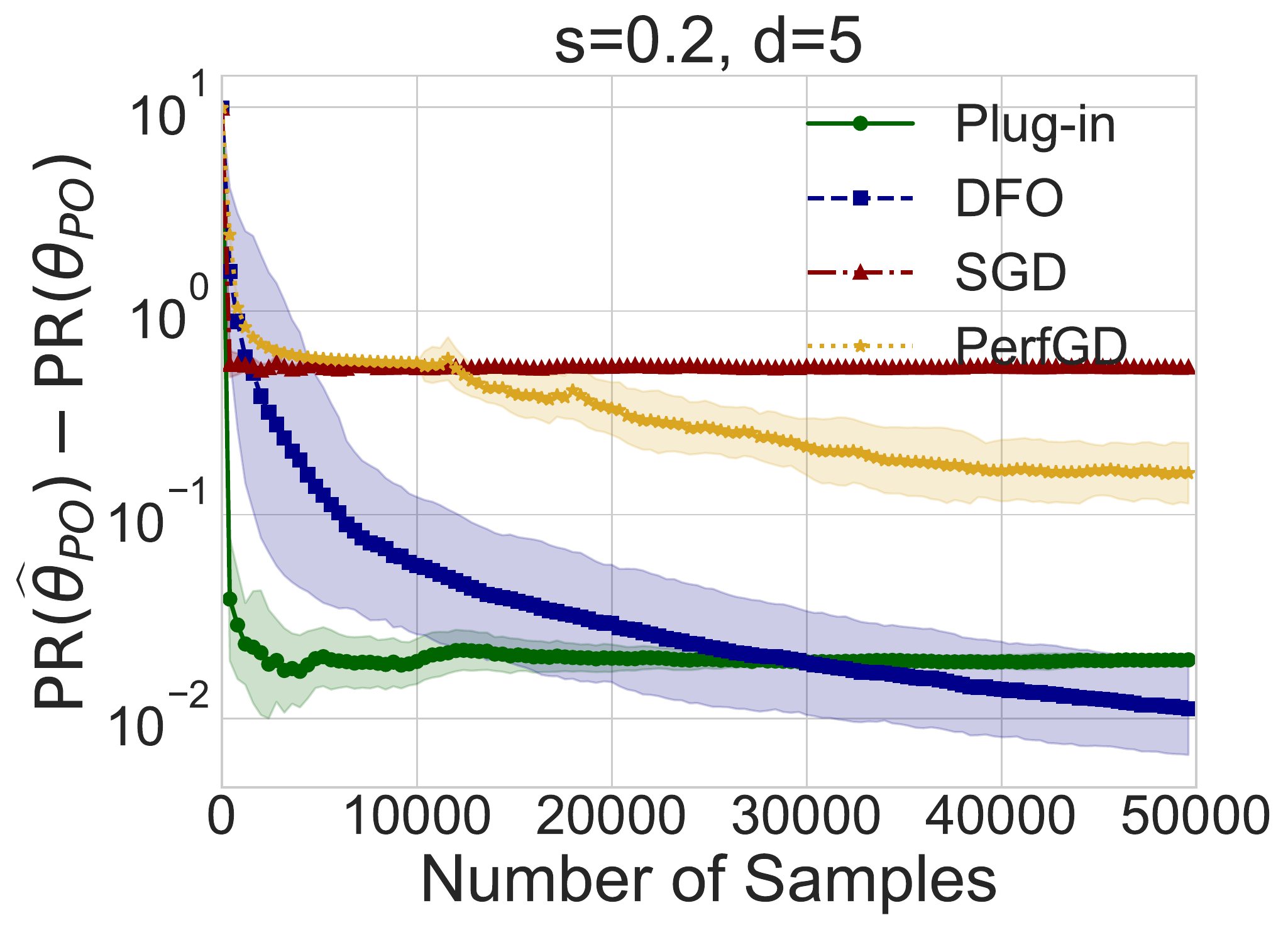}
    \includegraphics[width=0.3\textwidth]{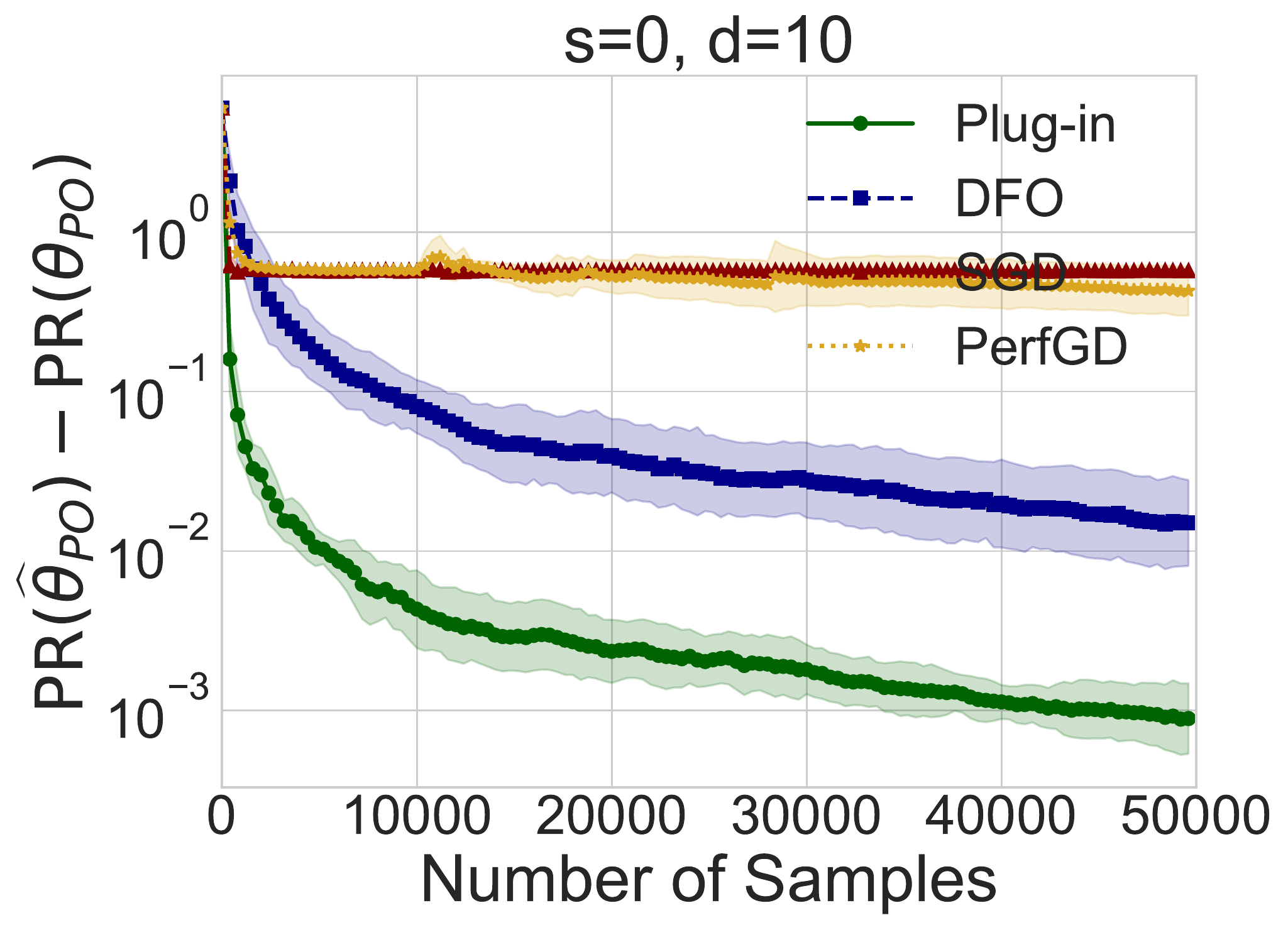}
   \includegraphics[width=0.3\textwidth]{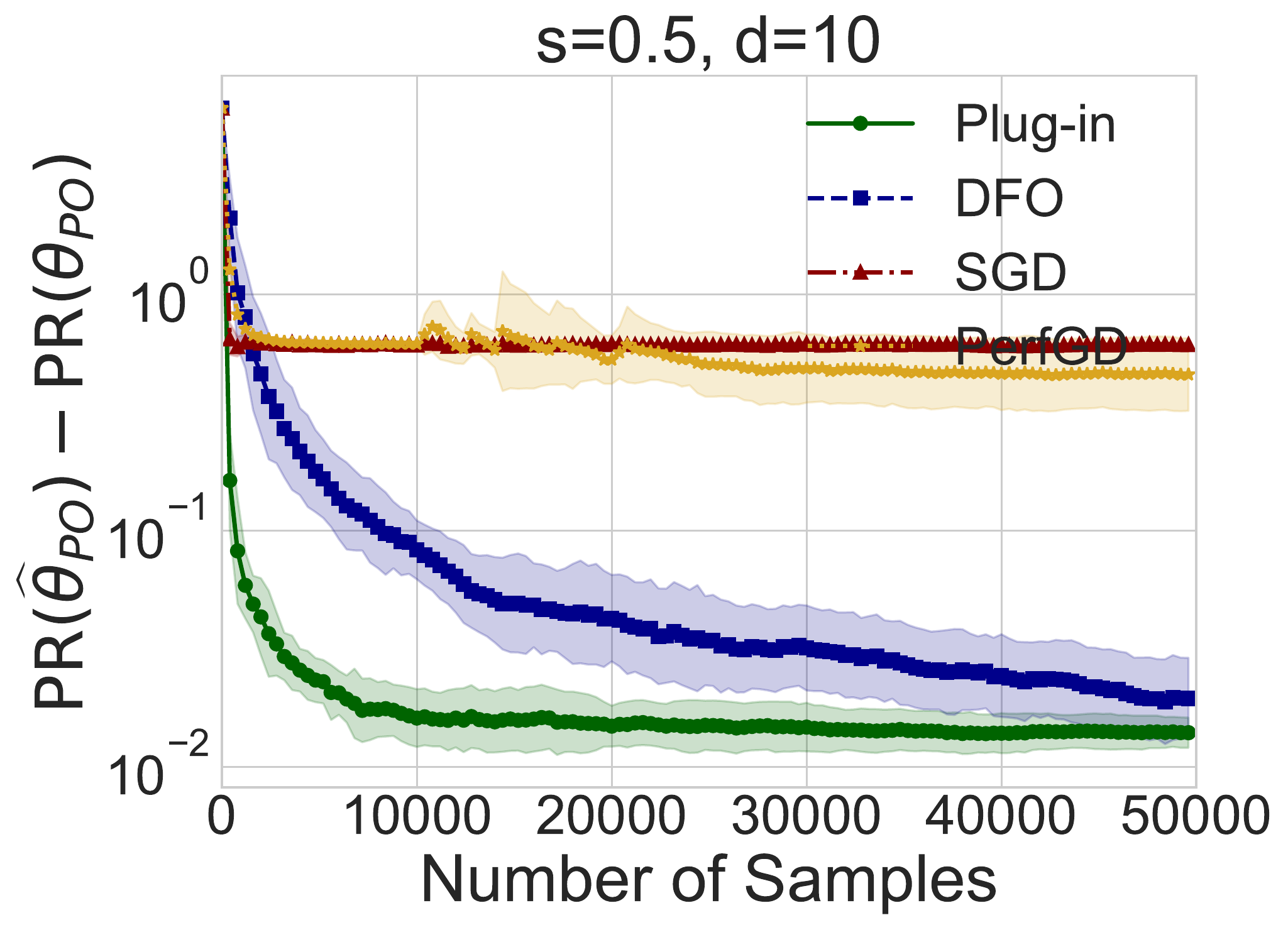}
  \includegraphics[width=0.3\textwidth]{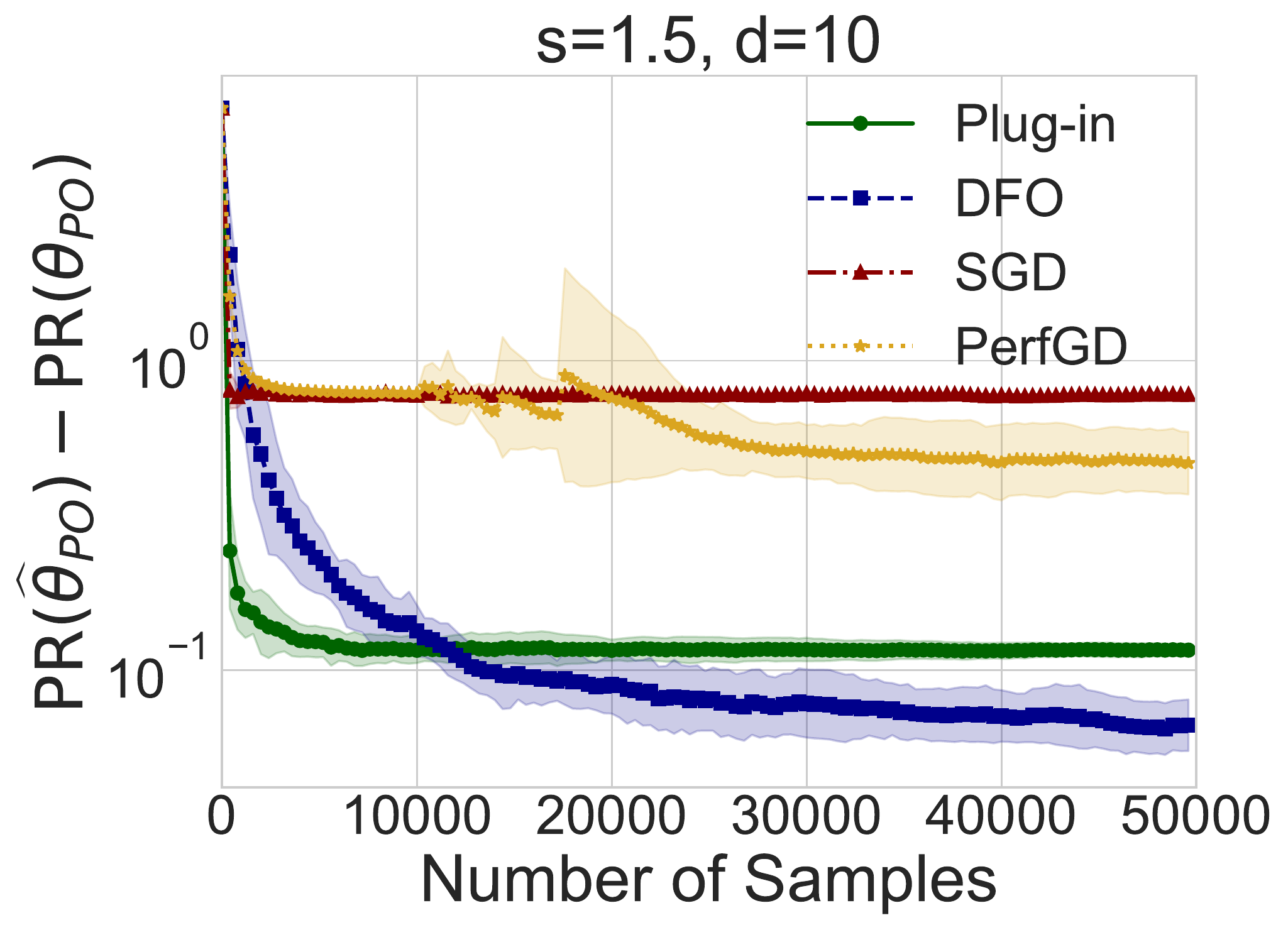}
  \caption{\textbf{(Location family)} Excess risk of plug-in performative optimization, DFO, greedy SGD, and PerfGD  with $\pm 1$ standard deviation on a logarithmic scale.}
  \label{fig:location_risk_compare_1}
\end{figure}

In Figure~\ref{fig:location_risk_compare_1} we see that the excess risk of our algorithm converges rapidly to a value that reflects the degree of misspecification. 
It approaches zero for $s=0$ (left panel) due to no misspecification and stabilizes at a nonzero value for $s>0$ (middle and right panels), consistent with our theory. In contrast, the risk of both PerfGD and DFO reduce slowly, while SGD quickly reaches a suboptimal value.

% 
% When $s=0$ (left panel), the excess risk of our algorithm approaches zero as the number of samples increases, given there is no misspecification.  On the other hand, for $s>0$ (middle and right panels), the excess risk of our algorithm converges to a nonzero value, in accordance with our theoretical guarantee.
% Meanwhile, we see that the excess risk of DFO converges to zero at a slow rate, while SGD quickly converges to a highly suboptimal point.

\begin{figure}[t]
  \centering
    \includegraphics[width=0.3\textwidth]{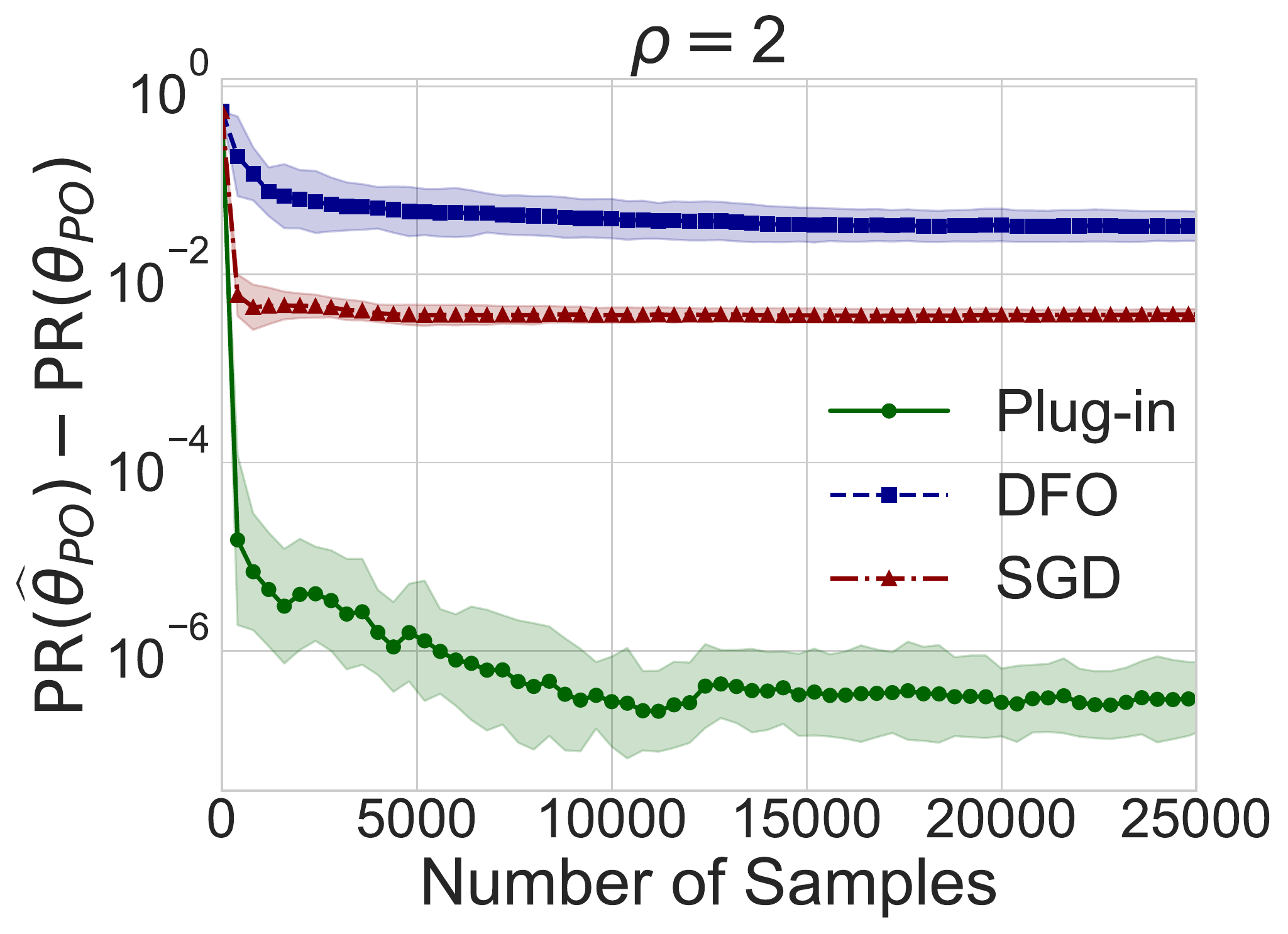}
      \includegraphics[width=0.3\textwidth]{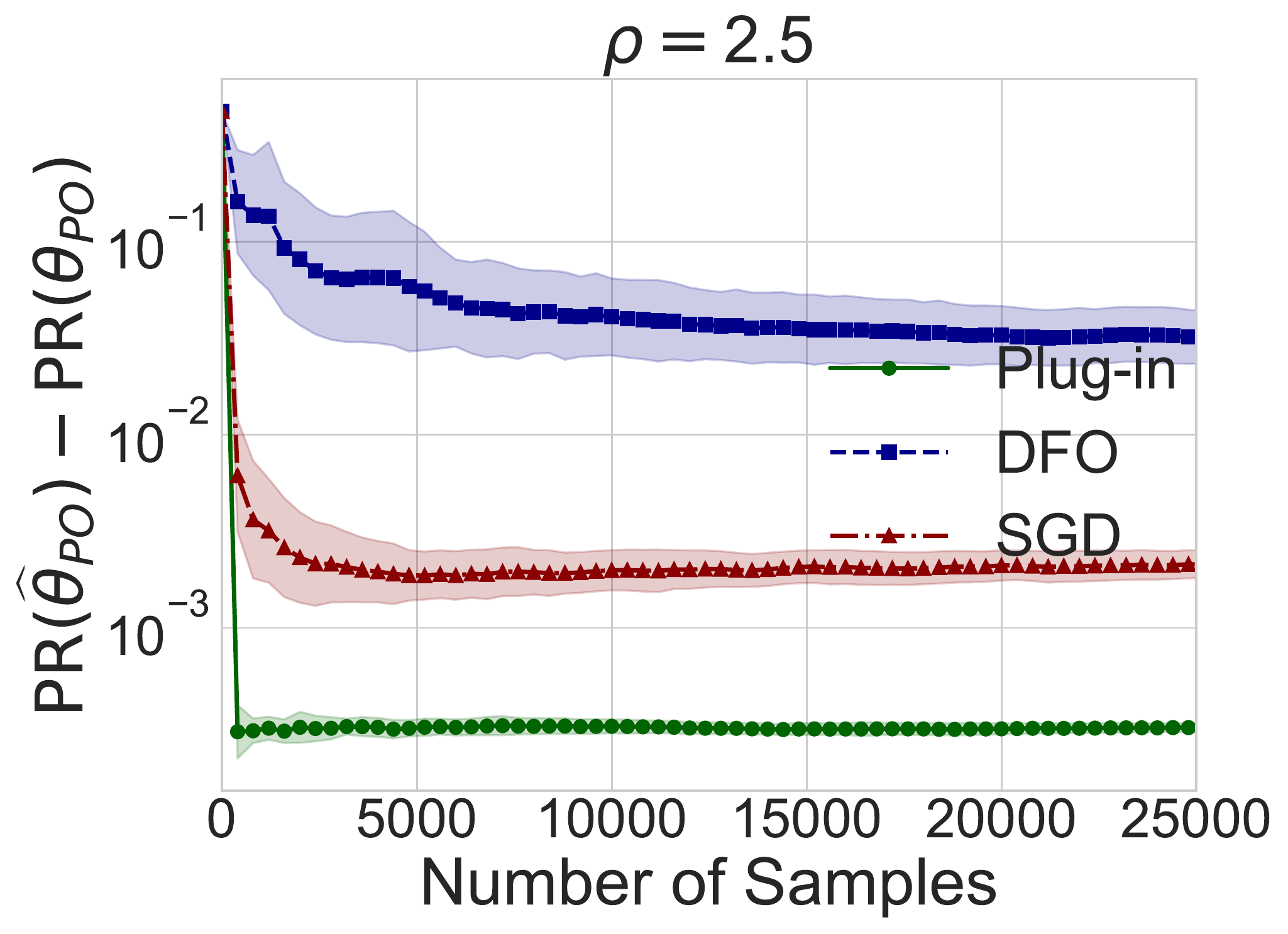}
  \includegraphics[width=0.3\textwidth]{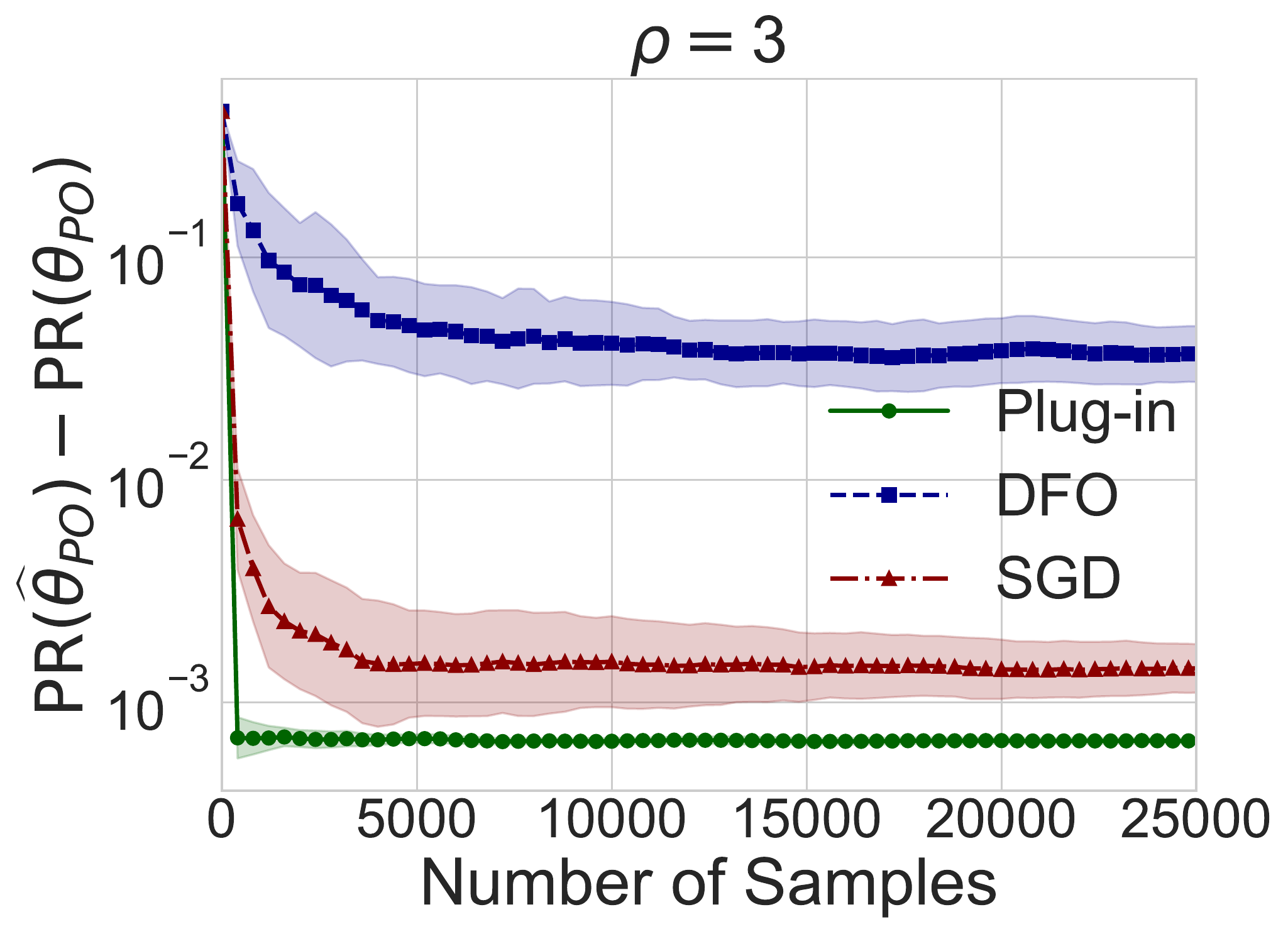}
  \includegraphics[width=0.3\textwidth]{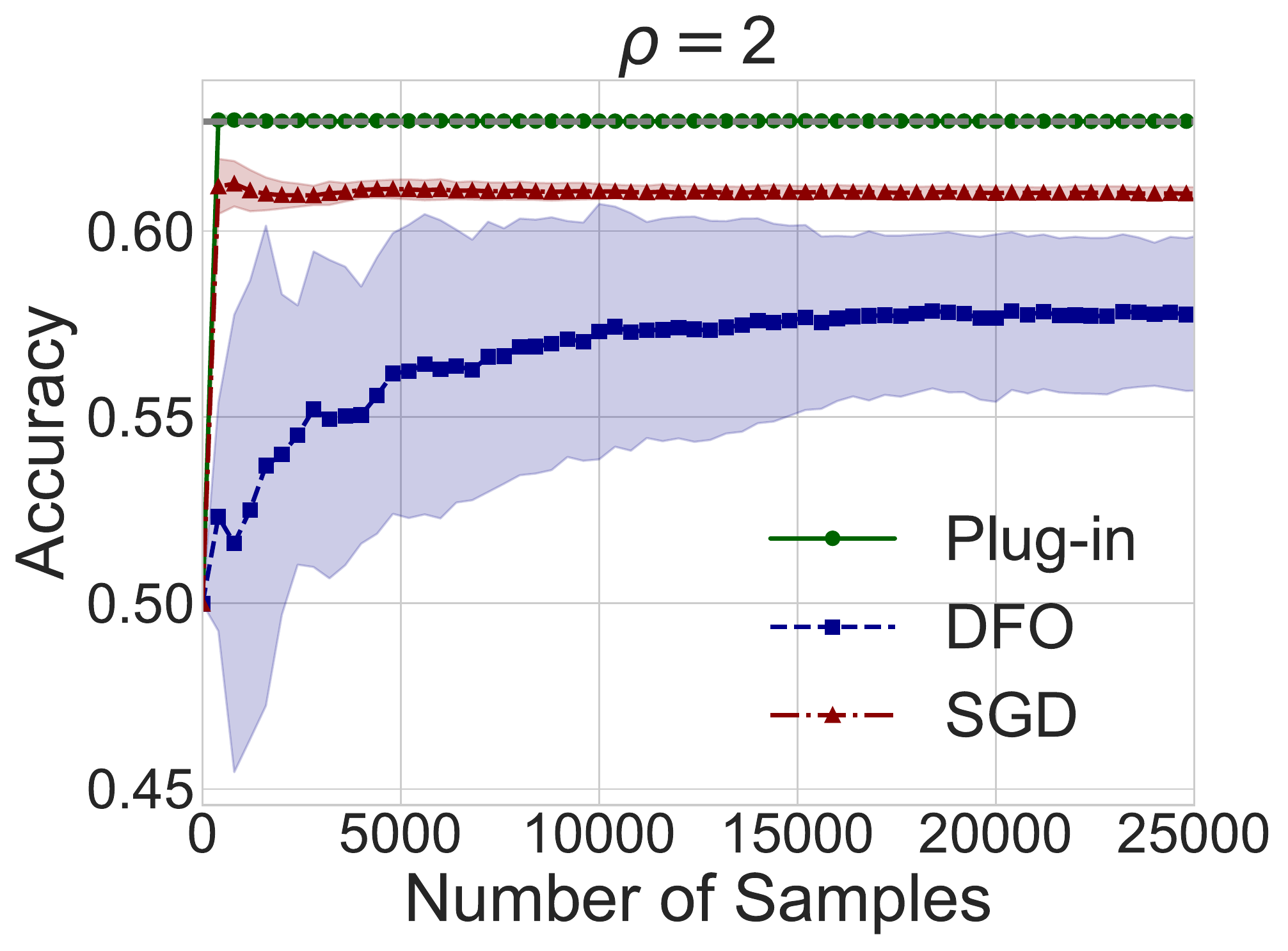}
  \includegraphics[width=0.3\textwidth]{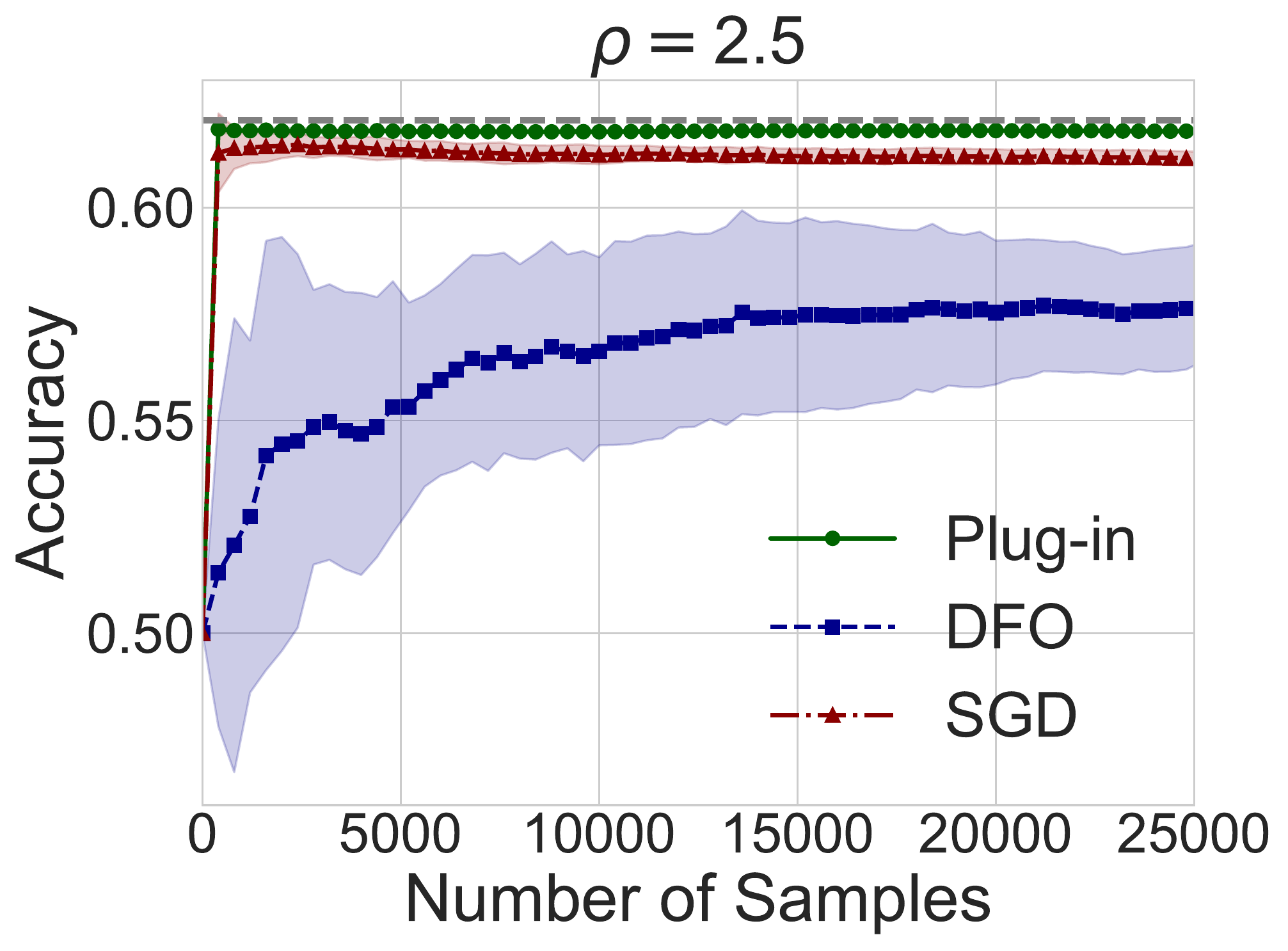}
  \includegraphics[width=0.3\textwidth]{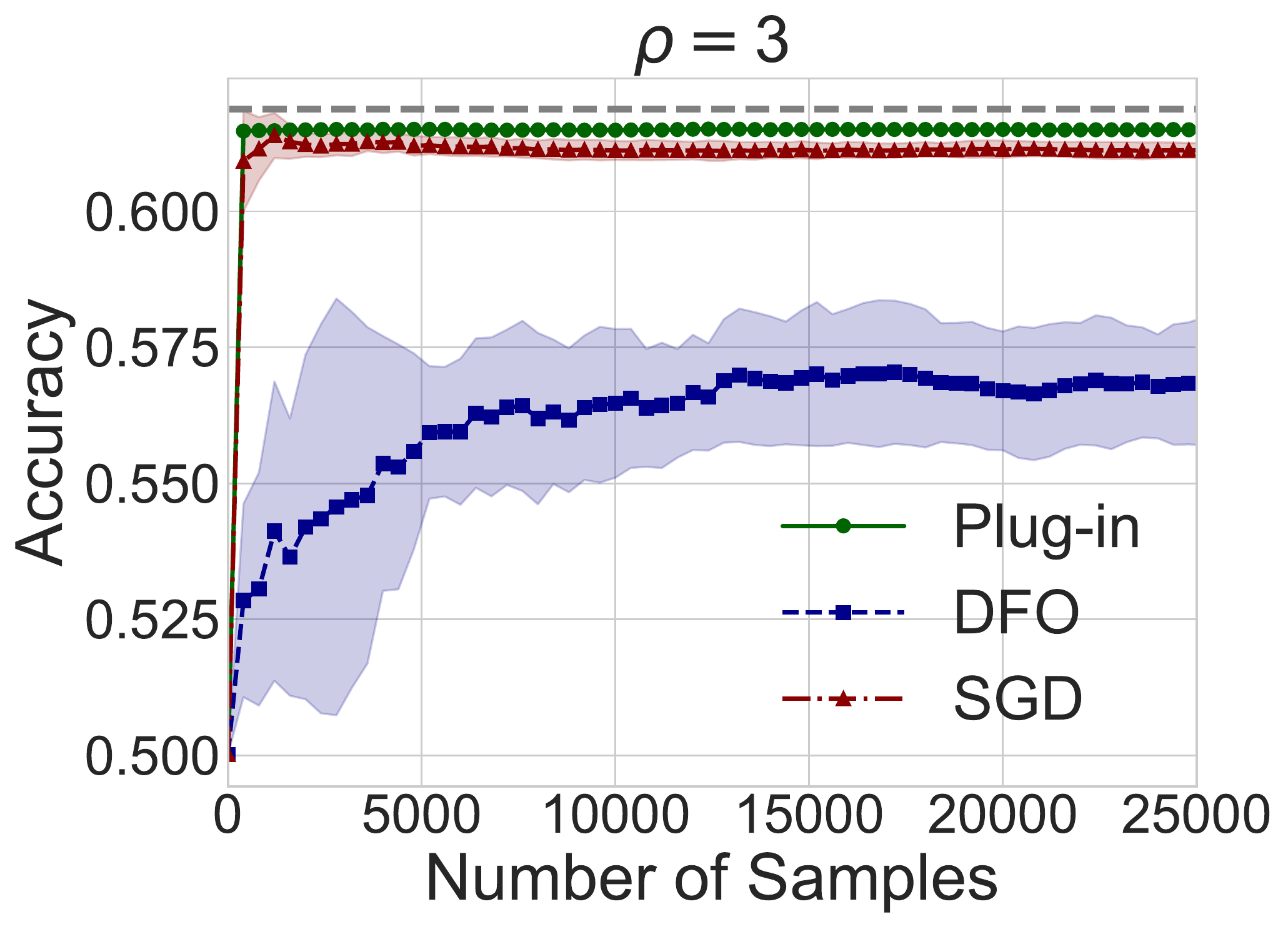}
  \caption{\textbf{(Strategic regression)} Excess risk and accuracy of plug-in performative optimization, DFO, and greedy SGD, with
 $\pm 1$ standard deviation on a logarithmic  scale.}
  \label{fig:strat_regr_compare_1}
\end{figure}

\subsection{Strategic regression}

We next consider strategic regression. We first generate $5000$ i.i.d. base samples  $(x_i,y_i)$ where $x_i\sim\N(0,\IdMat_{d_x})$, and $y_i\in\{0,1\}$ follows from a logistic model with a fixed parameter vector $\eta\sim\unif(\Sphere{d_x-1})$. Denote the joint empirical distribution of $(x_i,y_i)$ by $\D_{0}$. The true distribution map, $(x,y)\sim\D(\theta)$, is defined via
$$ (x_0,y_0)\sim\D_0, y=y_0, x = \argmax_{x'\in\R^{d_x}}(\theta^\top x'-\frac{1}{2\tilde\beta}\|x'-x_0\|_\rho^\rho),$$
for some $\rho>1$. We set  $d_x=5$ and $\tilde \beta=2$. To construct a model for $\D(\cdot)$, we follow the procedure from Section~\ref{sec:stra_regr}, using the linear utility and quadratic cost. Thus, $\rho=2$ results in correct specification. We choose $\ell(z;\theta)$ to be the logistic loss with a ridge penalty and $\Theta = \{\theta:\|\theta\|\leq 1\}$.
% \begin{align*}
%     \ell(x,y;\theta):=- y \log(1+\exp\{-x^\top\theta\})+(1-y)\log(1+\exp\{x^\top\theta\})+\lambda\enorm{\theta}^2,
% \end{align*} where the ridge parameter $\lambda=0.001$ and we restrict $\enorm{\theta}\leq 1$.

We examine the well-specified scenario $\rho=2$ and two misspecified cases $\rho\in\{2.5,3\}$, comparing our method with DFO and greedy SGD in Figure~\ref{fig:strat_regr_compare_1}. We see our algorithm quickly tends to zero excess risk when $\rho=2$ (top left).   When $\rho\neq 2$ (middle and right), the algorithm still converges, albeit to a suboptimal point. SGD converges to a suboptimal point with nonzero excess risk, while the excess risk of the DFO method decreases at a slow rate. We remark that PerfGD is not applicable in our strategic regression setting as $\D(\theta)$ does not admit a probability density function.  Similar trends appear for test accuracy.

% We consider both the well-specified scenario, $\rho=2$, and two misspecified scenarios, $\rho\in\{2.5,3\}$, and compare our method with DFO and greedy SGD. The results are shown in Figure~\ref{fig:strat_regr_compare_1}.
% %
% As before, our algorithm exhibits rapid convergence to a point with zero excess risk when the model is well-specified (top left). Under misspecification (top middle and right), the algorithm still converges, albeit to a somewhat suboptimal point. In contrast, SGD converges to a suboptimal point with nonzero excess risk, while the excess risk of the DFO method decreases at a slow rate.
% In addition, we also observe similar trends in terms of test accuracy (bottom panel). 

% It should be noted that in the misspecified scenarios $(\rho=2.5,3)$,  both our algorithm and SGD achieve a test accuracy close to  $\thetaPO$ due to the small magnitude of the misspecification error.

\section{Discussion}

We have analyzed performative prediction with misspecified models. Our results highlight the statistical gains of using models, however modeling has consequences far beyond statistical efficiency. On the positive side, modeling can help interpretability and computational efficiency. On the negative side, however, using highly misspecified models can lead to unfairness, lack of validity, and poor downstream decisions.  For example, modeling may be too coarse and not represent certain demographic groups properly.
Going forward, it would be valuable to develop deeper understanding of such negative aspects of modeling. Overall, given that models are ubiquitous in practice, we believe they merit further study—especially under misspecification—and we have only scratched the surface of this agenda.

We discuss several limitations and comment on a few technical aspects of our work. Along the way, we discuss valuable future directions in the context of modeling the distribution map in performative prediction.

First, our approach may be of limited use when the learner has very little prior knowledge of the true data-generating process, as this prevents them from choosing an atlas in an informed way. This is a hard scenario, and it is possible that no existing method would lead to a satisfactory performance. For example, while the model-free approach of Jagadeesan et al.~\cite{jagadeesan2022regret} is asymptotically more robust, for a reasonable number of samples it is not clear that it would outperform our approach with a highly misspecified model. 
One valuable future direction is to choose a family of distribution maps $\MapFam_{\B}$ that is sufficiently complex and has strong expressive power, e.g., neural networks, in cases where the learner is very uncertain about the true map. When the number of samples is suitably large, this may give a small excess risk as the misspecification error is smaller with a more expressive model. While in the current paper we mainly focus on simple models with closed-form expressions, we are hopeful that our approach could be valuable in such settings with black-box modeling as well.

Modeling could be particularly useful if one considers the fact that in reality data distributions take time to shift after model deployment and do not generate i.i.d. observations. Such time-varying shifts were, in fact, modeled in prior work \citep{brown2022performative,izzo2022learn,li2022state,ray2022decision}. It would be interesting to study the impacts of modeling the time-varying aspect of distribution maps.

Many of our examples assumed access to historical data. However, this assumption is not fundamental to our framework. Often this just means that there was another historical model in place under which the data was collected. We can model this as ``step 0'' in our framework, where we first deploy a ``default'' model $\theta_0$ (e.g. $\theta_0=0$) and collect data points drawn from $\mathcal{D}(\theta_0)\equiv \mathcal{D}_0$ (for instance, in Example \ref{ex:perf_outcomes} this would correspond to deploying a model with $f_\theta(x)=0$ for all $x$; there exist analogues for other problems as well). This alternative view would simply incur an additional statistical error term due to having finite-sample access to $\mathcal{D}(\theta_0)$.

In this work, we used standard ERM to estimate the distribution map. However, the learner may want to incorporate criteria other than model fit (which is the focus on ERM) in this process. For example, the learner may want to use a classical model selection criterion such as AIC or BIC (see, e.g.,~\cite{hastie2009elements}) to regularize towards ``simpler'' models and run ERM afterwards. Investigating model selection criteria beyond ERM for distribution map estimation is another valuable future direction.

\bibliographystyle{plain}
\bibliography{references}

\newpage

\appendix

\section{Proofs}

\subsection{Notation and definitions}

In the proofs we will sometimes use $c,c'>0$ to denote universal constants and $C,C'>0$ to denote constants that may depend on the parameters introduced in the assumptions. We allow the values of the constants to vary from place to place.

We say a random variable $x$ is \textit{subexponential} with parameter $\nu$ if \[\P\{|x|\geq t\}\leq 2\exp\left(-\frac{t}{\nu}\right) \] for any $t\geq 0$. Unless specified, we do not assume $x$ has mean zero in general. Moreover, we say a vector $\mathbf{x}\in\R^d$ is subexponential with parameter $\nu$ if $\inprod{\Direc}{\mathbf{x}}$ is subexponential with parameter $\nu$ for any fixed direction $\Direc\in\mathcal{S}^{d-1}$. Similarly, we say a random variable $x$ is \textit{subgaussian} with parameter $\sigma$ if \[\P\{|x|\geq t\}\leq 2\exp\left(-\frac{t^2}{\sigma^2}\right) \] for any $t\geq 0$. Likewise, a vector $\mathbf{x}\in\R^d$ is subgaussian with parameter $\sigma$ if $\inprod{\Direc}{\mathbf{x}}$ is subgaussian with parameter $\sigma$ for any fixed direction $\Direc\in\mathcal{S}^{d-1}$.

\subsection{Proof of Theorem \ref{thm:general_risk_bound}}

Define the population-level counterpart of $\thetahat$ as:
\[\theta^* = \argmin_{\theta\in\Theta} \PR^{\beta^*
}(\theta).\]

We can write
\begin{align*}
&\PR(\thetahat) - \PR(\thetaPO)\\
&\quad = (\PR(\thetahat) - \PR^{\beta^*}(\thetahat)) + (\PR^{\beta^*}(\thetahat)- \PR^{\hat\beta}(\thetahat)) + (\PR^{\hat\beta}(\thetahat) -\PR^{\hat\beta}(\theta^*))\\
&\quad + (\PR^{\hat\beta}(\theta^*) - \PR^{\beta^*}(\theta^*)) + (\PR^{\beta^*}(\theta^*) - \PR^{\beta^*}(\thetaPO)) + (\PR^{\beta^*}(\thetaPO)  - \PR(\thetaPO)).
\end{align*}
By the definition of $\thetahat$, we know $\PR^{\hat \beta}(\hat \theta)- \PR^{\hat \beta}(\theta^*) \leq 0$. Similarly, by the definition of $\theta^*$, we know $\PR^{\beta^*}(\theta^*)  - \PR^{\beta^*}(\thetaPO) \leq 0$. Using these inequalities, we establish
\begin{align*}
\PR(\thetahat) - \PR(\thetaPO) &\leq (\PR(\thetahat) - \PR^{\beta^*}(\thetahat)) + (\PR^{\beta^*}(\thetahat)- \PR^{\hat\beta}(\thetahat))\\
&\quad + (\PR^{\hat\beta}(\theta^*) - \PR^{\beta^*}(\theta^*))  + (\PR^{\beta^*}(\thetaPO)  - \PR(\thetaPO))\\
&\leq 2\sup_{\theta}|\PR(\theta) - \PR^{\beta^*}(\theta)| + 2\sup_{\theta}|\PR^{\beta^*}(\theta) - \PR^{\hat \beta}(\theta)|\\
&= 2(\textup{\texttt{StatErr}}_n + \textup{\misspecerr}).
\end{align*}

\subsection{Proof of Corollary \ref{cor:tv_main}}

We have
\begin{align*}
  |\PR^{\beta^*}(\theta)-\PR(\theta)|
    &=\left|\int \ell(z;\theta)(p_{\beta^*}(z;\theta)-p(z;\theta))dz\right|\\
      &\leq
     B_\ell\cdot \int |p_{\beta^*}(z;\theta)-p(z;\theta)|dz\\
    &=2 B_\ell\cdot\TV(\D_{\beta^*}(\theta),\D(\theta)).
\end{align*}
Therefore,
\[\misspecerr \leq 2 B_\ell \sup_{\theta\in\Theta}\TV(\D_{\beta^*}(\theta),\D(\theta)) \leq 2 B_\ell \eta_{\TV}.\]
By a similar argument as above, $|\PR^{\beta^*}(\theta)-\PR^{\hat \beta}(\theta)| \leq 2B_\ell \TV(\D_{\beta^*}(\theta),\D_{\hat \beta}(\theta))$. Applying $\epsilon_\TV$-smoothness of the distribution atlas, we get
\[\staterr \leq 2B_\ell \sup_{\theta\in\Theta} \TV(\D_{\beta^*}(\theta),\D_{\hat \beta}(\theta)) \leq 2B_\ell \epsilon_\TV \|\hat\beta - \beta^*\| \leq 2B_\ell \epsilon_\TV C_n.\]
Applying Theorem \ref{thm:general_risk_bound} completes the proof.

\subsection{Proof of Corollary \ref{cor:wass_main}}

Denote by $\couple(\D,\D')$ a coupling between two distributions $\D$ and $\D'$. We have
\begin{align*}
   \left|\PR^{\beta^*}(\theta)-\PR(\theta)\right|
    &=\left| \inf_{\Pi(\D_{\beta^*}(\theta),\D(\theta))} \E_{(z,z')\sim \couple(\D_{\beta^*}(\theta),\D(\theta))} [\ell(z;\theta)- \ell(z';\theta)]\right|\\
      &\leq
      \inf_{\couple(\D_{\beta^*}(\theta),\D(\theta))}\E_{(z,z')\sim \couple(\D_{\beta^*}(\theta),\D(\theta))} [|\ell(z;\theta)- \ell(z';\theta)|]\\
        &\leq
      L \inf_{\couple(\D_{\beta^*}(\theta),\D(\theta))}\E_{(z,z')\sim \couple(\D_{\beta^*}(\theta),\D(\theta))} [\|z-z'\|]\\
      &=L\mathcal W(\D_{\beta^*}(\theta),\D(\theta)).
\end{align*}
Therefore,
\[\misspecerr \leq L \sup_{\theta\in\Theta}\mathcal W(\D_{\beta^*}(\theta),\D(\theta)) \leq L\cdot \eta_W.\]
By a similar argument, $|\PR^{\beta^*}(\theta)-\PR^{\hat \beta}(\theta)| \leq L \mathcal W(\D_{\beta^*}(\theta),\D_{\hat \beta}(\theta))$. Applying $\epsilon_W$-smoothness of the distribution atlas, we get
\[\staterr \leq L \sup_{\theta\in\Theta} \mathcal W(\D_{\beta^*}(\theta),\D_{\hat \beta}(\theta)) \leq L \epsilon_W \|\hat\beta - \beta^*\| \leq L \cdot \epsilon_W \cdot C_n.\]
Applying Theorem \ref{thm:general_risk_bound} completes the proof.

\subsection{Proof of Lemma~\ref{lemma:estimation_lemma}}
We first present Lemma~\ref{lm:consistent_cvx} which we will use in the proof.
\begin{lemma}\label{lm:consistent_cvx}
    \begin{align}
\inprod{\nabla\MFun(\beta)}{\beta - \beta^*} & \geq
\begin{dcases}
 \frac{\mu}{2}\|{\beta^* - \beta}\|^2, &
 \|{\beta^* - \beta}\| \leq \frac{\mu}{\sigma_\MFun},\\
 \frac{\mu^2}{2\sigma_\MFun} \|{\beta^* - \beta}\|,
 &\|\beta^* - \beta\|\ge \frac{\mu}{\sigma_\MFun}.
\end{dcases}
\end{align}
\end{lemma}
See the proof of Lemma~\ref{lm:consistent_cvx} in Supplement~\ref{sec:pf_lm:consistent_cvx}

Let $B_\beta$ be the bound of the parameter set $\B$, i.e., $\|\beta\|\leq\B_\beta$ for any $\beta\in\B$. We will show Lemma~\ref{lemma:estimation_lemma} holds with some sufficiently large constants $C,C'>0$ that depend polynomially on $(\log|1+ B_\beta|,1/\mu,L_\MFun,B_\MFun,\sigma_\MFun)$.

Denote 
\[\MFun_n(\beta):=\frac{1}{n}\sum_{i=1}^n\MFun(\theta_i,z_i;\beta);~~~ \MFun(\beta):=\E_{\theta\sim\Dexp, z\sim\D(\theta)}[\MFun(\theta,z;\beta)].\]

We begin by claiming  the following result, which we will prove later. With probability over $1-\delta$
\begin{align}
\sup_{\beta\in\B}\enorm{\nabla_\beta \MFun_n(\beta)- \nabla_\beta\MFun(\beta)}
&\leq
C \sqrt{\log n}\sqrt{\frac{{ d_\beta+\log(1/\delta)}}{n}}.
\label{eq:emp_proce_nabla_m}
\end{align}
With this result at hand, it follows from Lemma~\ref{lm:consistent_cvx} that
\begin{align}
\min\Big\{\frac\mu2\|\hat\beta-\beta^*\|^2,\frac{\mu^2}{2\sigma_\MFun}\|\hat\beta-\beta^*\|\Big\}
&\leq
\inprod{\nabla_\beta \MFun(\hat\beta)}{\hat\beta-\beta^*}=\inprod{\nabla_\beta \MFun(\hat\beta)-\nabla_\beta \MFun_n(\hat\beta)}{\hat\beta-\beta^*}\notag\\
&\leq
\enorm{\nabla_\beta\MFun(\hat\beta)-\nabla_\beta \MFun_n(\hat\beta)}\enorm{\hat\beta-\beta^*}\leq C  \sqrt{\log n}\sqrt{\frac{{ d_\beta+\log(1/\delta)}}{n}}\enorm{\hat\beta-\beta^*},\label{eq:m_est_finite_major_ineq}
\end{align}
where the first equality is due to the fact that $\nabla_\beta\MFun_n(\hat\beta)=0$. 
Eliminating  $\enorm{\hat\beta-\beta^*}$ in both  the first and the last term of~\eqref{eq:m_est_finite_major_ineq} yields \begin{align*}
\frac{\mu^2}{2\sigma_{\MFun}}\wedge\frac{\mu}2\enorm{\hat\beta-\beta^*}\leq C\sqrt{\log n}\sqrt{\frac{{ d_\beta+\log(1/\delta)}}{n}}.
\end{align*}  By the sample size Assumption in Lemma~\ref{lemma:estimation_lemma}, we may assume $n$ is sufficiently large such that $\frac{\mu^2}{2\sigma_{\MFun}}\geq 
C\sqrt{\log n}\sqrt{\frac{{ d_\beta+\log(1/\delta)}}{n}}
$ for some constant $C>0$. Therefore, we conclude that
\begin{align*}
\enorm{\hat\beta-\beta^*}\leq C\sqrt{\log n}\sqrt{\frac{{ d_\beta+\log(1/\delta)}}{n}}.
\end{align*}

\paragraph{Proof of Eq.~\eqref{eq:emp_proce_nabla_m}.}
Let $\{\Direc_1,\Direc_2,\ldots,\Direc_M\}$ be a $1/2$-covering of $\Sphere{d_\beta-1}$ in the Euclidean norm such that $|M|\leq 5^{d_\beta}$. Define the random variables \[\ScoreFun_{\Direc,\beta}:=\inprod{\Direc}{\nabla_\beta \MFun_n(\beta)- \nabla_\beta\MFun(\beta)},
~~
\ScoreFun_\Direc:=\sup_{\beta\in\B}\ScoreFun_{\Direc,\beta}.
\] It follows from a standard discretization argument (e.g.,~\cite{wainwright2019high}, Chap. 6) that
\begin{align}
\sup_{\beta\in\B}\enorm{\nabla_\beta \MFun_n(\beta)- \nabla_\beta\MFun(\beta)}
\leq 
2\sup_{\beta\in\B}\sup_{i\in[M]}\ScoreFun_{\Direc_i,\beta}\label{eq:m_est_2norm_disc}.
\end{align}
We make the following claim which will be proved at the end. With probability over $1-\delta$
\begin{align}
\opnorm{\frac{1}{n}\sum_{i=1}^n\sup_{\beta\in\B} \nabla^2_{\beta}\MFun(\theta_i,z_i;\beta)}\leq c L_\MFun + cL_\MFun\sqrt{\frac{d_\beta+\log(1/\delta)}{n}}\leq cL_\MFun\label{eq:claim:general_hess}.
\end{align}
for some constant $c>0$.

Let $\eps>0$ be some value we specify later.
Construct an $\varepsilon$-covering net $\{\beta^1,\ldots,\beta^N\}$ of $\B$ in $\enorm{\cdot}$. Then the covering number $|N|\leq (1+2B_\beta/\eps)^{d_\beta}$, and 
\begin{align}
\sup_{\beta\in\B}\sup_{i\in[M]}\ScoreFun_{\Direc_i,\beta}
&\leq
\sup_{i\in[M]}\sup_{j\in[N]}\ScoreFun_{\Direc_i,\beta^j}+\sup_{i\in[M]}\sup_{\enorm{\beta_1-\beta_2}\leq\varepsilon}|\ScoreFun_{\Direc_i,\beta_1}-\ScoreFun_{\Direc_i,\beta_2}|\notag\\
&\leq
\sup_{i\in[M]}\sup_{j\in[N]}\ScoreFun_{\Direc_i,\beta^j}+
\sup_{i\in[M]}
\sup_{\enorm{\beta_1-\beta_2}\leq\varepsilon} \frac{1}{n}\sum_{i=1}^n\sup_{\beta\in\B}\Direc_i^\top \nabla^2_{\beta}\MFun(\theta_i,z_i;\beta)(\beta_1-\beta_2)\notag\\
&\leq
\sup_{i\in[M]}\sup_{j\in[N]}\ScoreFun_{\Direc_i,\beta^j}+
\sup_{i\in[M]}
\sup_{\enorm{\beta_1-\beta_2}\leq\varepsilon}
\enorm{\Direc_i}\opnorm{\frac{1}{n}\sum_{i=1}^n\sup_{\beta\in\B} \nabla^2_{\beta}\MFun(\theta_i,z_i;\beta)}\enorm{\beta_1-\beta_2}
\notag\\
&\leq
\sup_{i\in[M]}\sup_{j\in[N]}\ScoreFun_{\Direc_i,\beta^j}+
cL_\MFun\varepsilon
,\label{eq:finite_sam_discre}
\end{align}
where the last inequality follows from the claim in ~\eqref{eq:claim:general_hess}.
Since \mbox{$\inprod{\Direc}{\nabla_\beta\MFun(\theta_i,z_i;\beta)-\E[\nabla_\beta\MFun(\theta_i,z_i;\beta)]}$} is zero mean subexponential with parameter $B_\MFun$ by condition \ref{assn:m_est_finite_bnd} of Assumption~\ref{ass:regularity}, it follows from concentration of subexponential variables that
\[
\P\{|\ScoreFun_{u_i,\beta^j}|\geq  t\}
\leq 
2 \exp\left(-\frac{nt^2}{2 B_\MFun^2}\right)~~\text{ for any } |t|\leq B_\MFun.\] Applying a union bound over $i,j$, we establish
\begin{align}
\P\left\{\max_{i\in[M],j\in[N]}|\ScoreFun_{u_i,\beta^j}|\geq t\right\}\leq  2 \exp\left(cd_\beta(\log(1+2B_\beta/\eps)+1)-\frac{nt^2}{2 B_\MFun^2}\right)~~\text{ for any } |t|\leq B_\MFun.\label{eq:m_est_finite_union}
\end{align}
Let $\eps=\sqrt{\frac{d_\beta+\log(1/\delta)}{n}}$ and 
\begin{align*}
t=\frac{cB_\MFun\sqrt{\log(1/\delta)+d_\beta(\log(1+2B_\beta/\eps)+1)}}{\sqrt{n}}
&\leq 
\frac{C B_\MFun\sqrt{\log n (d_\beta+\log(1/\delta))}}{\sqrt{n}}\leq {B_\MFun}
\end{align*} for some constant $c>0$,
where the last inequality uses the sample size assumption of the lemma.
Substituting the values of $\eps$ and $t$ into Equations~\eqref{eq:finite_sam_discre},~\eqref{eq:m_est_finite_union} and combining with Eq.~\eqref{eq:m_est_2norm_disc}, we obtain
\begin{align*}
\sup_{\beta\in\B}\enorm{\nabla_\beta \MFun_n(\beta)- \nabla_\beta\MFun(\beta)}
&\leq
\frac{C B_\MFun\sqrt{\log n (d_\beta+\log(1/\delta))}}{\sqrt{n}}+cL_\MFun \sqrt{\frac{d_\beta+\log(1/\delta)}{n}}\\
&\leq
C'  \sqrt{\log n}\sqrt{\frac{{ d_\beta +\log(1/\delta)}}{n}}
\end{align*}
with probability over $1-\delta$ for some parameter-dependent constant $C'$.

\paragraph{Proof of Eq.~\eqref{eq:claim:general_hess}.}
Similar to equation~\eqref{eq:m_est_2norm_disc}, from a standard discretization argument we have
\begin{align*}
\opnorm{\frac{1}{n}\sum_{i=1}^n\sup_{\beta\in\B} \nabla^2_{\beta}\MFun(\theta_i,z_i;\beta)}
\leq 
2\sup_{j\in[M]}\frac{1}{n}\sum_{i=1}^n\sup_{\beta\in\B} \Direc_j^\top\nabla^2_{\beta}\MFun(\theta_i,z_i;\beta)\Direc_j.
\end{align*}
Since $\sup_{\beta\in\B}\Direc_j^\top \nabla^2_{\beta}\MFun(\theta_i,z_i;\beta)\Direc_j$ are subexponential variables by condition \ref{assn:m_est_finite_lip} in Assumption \ref{ass:regularity}, we have from properties of subexponential variables and Bernstein's inequality  that 
\begin{align*}
&\quad \P\left\{\Direc_j^\top\frac{1}{n}\sum_{i=1}^n\sup_{\beta\in\B} \nabla^2_{\beta}\MFun(\theta_i,z_i;\beta)\Direc_j\geq c L_\MFun + t\right\}\\
&\leq
\P\left\{\Direc_j^\top\frac{1}{n}\sum_{i=1}^n\sup_{\beta\in\B} \nabla^2_{\beta}\MFun(\theta_i,z_i;\beta)\Direc_j\geq  \E[\Direc_j^\top\sup_{\beta\in\B} \nabla^2_{\beta}\MFun(\theta_i,z_i;\beta)\Direc_j] + t\right\}\\
&\leq \exp\left(-c\min\{\frac{nt}{L_\MFun},\frac{nt^2}{L_\MFun^2}\}\right).
\end{align*}
Applying a union bound over $j\in[M]$ and setting $t=cL_\MFun\sqrt{\frac{d_\beta+\log(1/\delta)}{n}}<c L_\MFun$, we establish
\begin{align*}
    \Direc_j^\top\frac{1}{n}\sum_{i=1}^n\sup_{\beta\in\B} \nabla^2_{\beta}\MFun(\theta_i,z_i;\beta)\Direc_j\leq c L_\MFun + cL_\MFun\sqrt{\frac{d_\beta+\log(1/\delta)}{n}}\leq cL_\MFun
\end{align*}
for some $c>0$ with probability over $1-\delta$.

\subsection{Proof of Lemma~\ref{lm:consistent_cvx}}\label{sec:pf_lm:consistent_cvx}
For any $ \|{\beta^* - \beta}\| \leq \frac{\mu}{\sigma_\MFun}$, by a Taylor expansion of $\nabla\MFun(\beta)$  at $\beta^*$ and Ass.~\ref{ass:regularity}(a), we have 
\begin{align*}
   \inprod{\nabla\MFun(\beta)}{\beta - \beta^*}=\inprod{\nabla\MFun(\beta)-\nabla\MFun(\beta^*)}{\beta - \beta^*} \geq{\mu}\|\beta-\beta^*\|^2-\frac{\sigma_\MFun}{2}\|\beta^*-\beta\|^3\geq\frac{\mu}{2}\|\beta^* - \beta\|^2.
\end{align*}
This gives the second part of Lemma~\ref{lm:consistent_cvx}.
When $\|{\beta^* - \beta}\| \geq \frac{\mu}{\sigma_\MFun}$, write  $\beta=\beta^*+t\Direc$, where $\Direc=(\beta-\beta^*)/\|\beta-\beta^*\|$. For a fixed direction $\Direc$, define $\beta(t):=\beta^*+t\Direc$ and
\begin{align*}
     f(\Direc,t):=\Big\langle{\nabla\MFun(\beta(t))},\frac{\beta(t) - \beta^*}{\|\beta(t) - \beta^*\|}\Big\rangle=\inprod{\Direc}{\nabla\MFun(\beta(t))}.
\end{align*}
Then for $t\geq0$, using Ass.~\ref{ass:regularity}(a) again we obtain $$
\partial_t f(\Direc,t)=\inprod{\Direc}{\nabla^2\MFun(\beta(t))\Direc}\geq0
$$
Therefore $f(\Direc,t)$ is increasing in $t$ and for any $\|\beta-\beta^*\|\geq\frac{\mu}{\sigma_{\MFun}}$
\begin{align*}
\Big\langle{\nabla\MFun(\beta)},{\frac{\beta - \beta^*}{\|\beta - \beta^*\|}}\Big\rangle&\geq  \Big\langle\nabla\MFun\Big(\beta\Big(\frac{\mu}{\sigma_\MFun}\Big)\Big),\frac{\beta\Big(\frac{\mu}{\sigma_\MFun}\Big) - \beta^*}{\|\beta\Big(\frac{\mu}{\sigma_\MFun}\Big) - \beta^*\|}\Big\rangle\\
&\geq
\frac{\mu}2\|\beta\Big(\frac{\mu}{\sigma_\MFun}\Big) - \beta^*\|
=
\frac{\mu^2}{2\sigma_\MFun}.
\end{align*}
This gives the second part of Lemma~\ref{lm:consistent_cvx}.

\subsection{Proof of Theorem \ref{thm:risk_bound_w_est_rate}}

The first statement follows directly by putting together Corollary \ref{cor:tv_main} and Lemma \ref{lemma:estimation_lemma}. Similarly, the second statement follows by putting together Corollary \ref{cor:wass_main} and Lemma \ref{lemma:estimation_lemma}.

\subsection{Proof of Claim \ref{claim:strat_reg_smoothness}}

Fix $x$ and $\theta$. We will use the shorthand notation $g_\beta(x,\theta)\equiv x_\beta$. First, we show that $\|x_{\beta}-x_{\beta'}\|_2 \leq \frac{B_u}{1-\beta_{\max} L_u} |\beta - \beta'|$.
To see this, notice that the optimality condition of the best-response equation is equal to:
\[ \beta \cdot \nabla u_\theta(x_\beta) -  x_\beta = -  x.\]
Since $\beta \nabla u_\theta(x_\beta) -  x_\beta = \beta' \nabla u_\theta(x_{\beta'}) -  x_{\beta'} =  -  x$, we know
\[\left\|\beta \nabla u_\theta(x_\beta) - \beta' \nabla u_\theta(x_{\beta'}) \right\| = \|x_{\beta} - x_{\beta'}\|.\]
We also know
\begin{align*}
\left\|\beta \nabla u_\theta(x_\beta) - \beta' \nabla u_\theta(x_{\beta'})\right\| &= \left\|\beta \nabla u_\theta(x_\beta) - \beta \nabla u_\theta(x_{\beta'}) + \beta \nabla u_\theta(x_{\beta'}) - \beta' \nabla u_\theta(x_{\beta'})\right\|\\
&\leq \beta L_u \|x_\beta - x_{\beta'}\| + B_u \left|\beta - \beta'\right|.
\end{align*}

Therefore,
\[\|x_\beta - x_{\beta'}\| \leq \beta L_u \|x_\beta - x_{\beta'}\| + B_u \left|\beta - \beta'\right|.\]
Rearranging the terms, we get
\[\|x_\beta - x_{\beta'}\| \leq \frac{B_u}{1-\beta L_u}|\beta - \beta'|.\]
By the definition of Wasserstein distance, this condition directly implies
\[\mathcal W(\D_{\beta}(\theta), \D_{\beta'}(\theta)) \leq \frac{B_u}{1-\beta_{\max} L_u}|\beta - \beta'|,\]
which is the definition of $\frac{B_u}{1-\beta_{\max} L_u}$-smoothness.

\subsection{Proof of Claim \ref{claim:strat_reg_rate}}

The claim follows by Lemma \ref{lemma:estimation_lemma} after verifying the conditions required in Assumption \ref{ass:regularity}. We have $r(\theta,x;\beta) =  \| x - \beta\nabla u_{\theta}(x)\|^2$, so $\nabla r(\theta,x;\beta) = -2(x^\top \nabla u_{\theta}(x) - \beta\|\nabla u_{\theta}(x)\|^2)$ and $\nabla^2 r(\theta,x;\beta) = 2\|\nabla u_{\theta}(x)\|^2$. Conditions (b) and (c) of Assumption \ref{ass:regularity} are thus satisfied by $\tilde x$ and $\nabla u_{\tilde \theta}(\tilde x)$ being subgaussian since products of subgaussians are subexponential. Condition (a) is satisfied by the fact that $r(\beta) = \E [\| \tilde x - \beta\nabla u_{\tilde \theta}(\tilde x)\|^2]$ is a quadratic in $\beta$ when $\E[\|\nabla u_{\tilde \theta}(\tilde x)\|^2]>0$.

\subsection{Proof of Claim \ref{claim:strat_class_smoothness}}

Fix $\theta,\beta,\beta'$, and without loss of generality let $\beta>\beta'$. We show that $\TV(\D_\beta(\theta),\D_{\beta'}(\theta))\leq \phi_u$. The distributions $\D_\beta(\theta)$ and $\D_{\beta'}(\theta)$ are equal to each other and to $\D_0$ for all $\{x:x^\top \theta \in (-\infty,T-\beta)\cup (T,\infty)\}$. Moreover, under both $\D_\beta(\theta)$ and $\D_{\beta'}(\theta)$, there is no mass for $\{x:x^\top \theta \in (T-\beta',T)\}$. The distributions thus only differ for $\{x:x^\top\theta \in[T-\beta,T-\beta']\cup\{T\}\}$. Since the density of $x^\top \theta$ is bounded by $\phi_u$, the measure of such vectors $x$ is at most $\phi_u|\beta-\beta'|$.

\subsection{Proof of Claim \ref{claim:strat_class_rate}}

By Hoeffding's inequality, with probability $1-\delta$ it holds that
\[\left|\frac 1 n \sum_{i=1}^n \mathbf{1}\{x_i^\top\theta_i \in (T\pm\epsilon)\} - \P\{\tilde x^\top\tilde \theta\in (T\pm\epsilon)\}\right| \leq \sqrt{\frac{\log(2/\delta)}{2n}}.\]
Let $\Delta = \sqrt{\frac{\log(2/\delta)}{2n}}$. Next we argue that $|\hat \beta - \beta^*|\leq \frac{\Delta}{\phi_l}$ by contradiction. Suppose $|\hat \beta - \beta^*|> \frac{\Delta}{\phi_l}$. Then,
\begin{align*}
\left|\frac 1 n \sum_{i=1}^n \mathbf{1}\{x_i^\top\theta_i \in (T\pm\epsilon)\} - \P\{\tilde x^\top\tilde \theta\in (T\pm\epsilon)\}\right| &= \left|\P\{x_0^\top\tilde \theta \in [T-\hat \beta,T] - \P\{x_0^\top\tilde \theta \in [T-\beta^*,T] \}\right|\\
&> \phi_l \frac{\Delta}{\phi_l}\\
&= \Delta,
\end{align*}
which contradicts Hoeffding's inequality. Therefore, we conclude that $|\hat \beta - \beta^*|\leq \frac{\Delta}{\phi_l}$.

\subsection{Proof of Claim~\ref{claim:loc_family_1}}

By the definition of Wasserstein distance, we have
\begin{align*}
 \mathcal W(\D_{\LinM_1}(\theta),\D_{\LinM_2}(\theta))
 =
 \mathcal W(\LinM_1\theta+z_0,\LinM_2\theta+z_0)=
 \enorm{\LinM_1\theta-\LinM_2\theta}\leq B_\theta\opnorm{\LinM_1-\LinM_2}
\end{align*}
for any $\LinM_1,\LinM_2$. Therefore,   the distribution atlas $\{\D_\LinM\}_{\LinM}$ is $\eps_W$-smooth with parameter $B_\theta$.

\subsection{Proof of Claim~\ref{claim:loc_family_2}}

Let $\subgtheta$ be the subgaussian parameter of $\Dexp$. We prove that there exists $C,C'$ depending polynomially on $(1/\kappa_{\min},\kappa_{\max},\subgtheta,L_{\theta z},B)$ such that Claim~\ref{claim:loc_family_2} holds.
By definition, we have
\begin{align*}
    \widehat\LinM^\top
    &=\left(\frac{1}{n}\sum_{i=1}^n \theta_i\theta_i^\top\right)^{-1}\Big(\frac{1}{n}\sum_{i=1}^n\theta_iz_{i}^\top\Big),\\
 \LinM^{*\top}
 &=
 \E[\tilde \theta\tilde \theta^\top]^{-1}\E[\tilde\theta\tilde z^\top].
\end{align*}
We state the  following results, which we  will  prove later:
\begin{align}
   \opnorm{\left(\frac{1}{n}\sum_{i=1}^n \theta_i\theta_i^\top\right)^{-1}-\E[\theta_i\theta_i^\top]^{-1}}
    &\leq
     \frac{c\subgtheta^2}{\kappa_{\min}^2}\sqrt{\frac{d_\theta +\log(1/\delta)}{n}}\leq C;
     \label{eq:linear_model_claim1}\\
    \opnorm{\frac{1}{n}\sum_{i=1}^n\theta_iz_{i}^\top-\E[\theta_iz_i^\top]}
    &\leq 
cL_{\theta,z}\sqrt{\frac{d_\theta+d_z+\log(1/\delta)}{n}}.
    \label{eq:linear_model_claim2}
\end{align}

Combining Equations~\eqref{eq:linear_model_claim1},~\eqref{eq:linear_model_claim2} with the assumptions of the claim, we establish
\begin{align*}
\opnorm{\widehat\LinM-\LinM^*}
&=
\opnorm{\left(\frac{1}{n}\sum_{i=1}^n \theta_i\theta_i^\top\right)^{-1}\Big(\frac{1}{n}\sum_{i=1}^n\theta_iz_{i}^\top\Big)
-\E[\theta_i\theta_i^\top]^{-1}\E[\theta_i z_i^\top]}\\
 &\leq
 \opnorm{\left(\frac{1}{n}\sum_{i=1}^n \theta_i\theta_i^\top\right)^{-1}-\E[\theta_i\theta_i^\top]^{-1}}\opnorm{\frac{1}{n}\sum_{i=1}^n\theta_iz_{i}^\top}
\\
& \quad+
\opnorm{\E[\theta_i\theta_i^\top]^{-1}}\opnorm{\frac{1}{n}\sum_{i=1}^n\theta_iz_{i}^\top-\E[\theta_iz_i^\top]}\\
&\leq C'\sqrt{\frac{d_\theta+d_z+\log(1/\delta)}{n}}.
\end{align*}
for some $C'>0$ that depends on problem-specific parameters.

\paragraph{Proof of Eq.~\eqref{eq:linear_model_claim1}.}
Under the conditions of the claim, we establish from concentration inequalities for subgaussian vectors (see, e.g., Theorem 6.5 in~\cite{wainwright2019high}) that with probability at least $1-\delta$,
\begin{align*}
\frac{\kappa_{\min}}{2}
&\leq \kappa_{\min}-c\subgtheta^2\sqrt{\frac{d_\theta +\log(1/\delta)}{n}}\leq \sigma_{\min}\left(\frac{1}{n}\sum_{i=1}^n \theta_i\theta_i^\top\right),\\
&\leq 
 \sigma_{\max}\left(\frac{1}{n}\sum_{i=1}^n \theta_i\theta_i^\top\right)
 \leq 
 \kappa_{\max}+
 c\subgtheta^2\sqrt{\frac{d_\theta +\log(1/\delta)}{n}}\leq \frac{3}{2}\kappa_{\max},
\end{align*}
where the last line follows from the sample-size assumption.
In addition, we also have from~\cite{wainwright2019high} that
\begin{align*}
    \opnorm{\frac{1}{n}\sum_{i=1}^n \theta_i\theta_i^\top-\E[\theta_i\theta_i^\top]}\leq c\subgtheta^2\sqrt{\frac{d_\theta +\log(1/\delta)}{n}}.
\end{align*}
Therefore, it follows from Woodbury's matrix identity and the last two displays that
\begin{align*}
     \opnorm{\left(\frac{1}{n}\sum_{i=1}^n \theta_i\theta_i^\top\right)^{-1}-\E[\theta_i\theta_i^\top]^{-1}}
     &=
     \opnorm{\left(\frac{1}{n}\sum_{i=1}^n \theta_i\theta_i^\top\right)^{-1}\left(\left(\frac{1}{n}\sum_{i=1}^n \theta_i\theta_i^\top\right)-\E[\theta_i\theta_i^\top]\right)(\E[\theta_i\theta_i^\top])^{-1}}\\
     &\leq 
     \opnorm{\left(\frac{1}{n}\sum_{i=1}^n \theta_i\theta_i^\top\right)^{-1}}\opnorm{\frac{1}{n}\sum_{i=1}^n \theta_i\theta_i^\top-\E[\theta_i\theta_i^\top]}\opnorm{(\E[\theta_i\theta_i^\top])^{-1}}\\
     &\leq 
     \frac{c\subgtheta^2}{\kappa_{\min}^2}\sqrt{\frac{d_\theta+\log(1/\delta)}{n}}.
\end{align*}
The second inequality follows from the assumption on sample size.

\paragraph{Proof of Eq.~\eqref{eq:linear_model_claim2}.}
Let $\{\Direc_1,\ldots,\Direc_M\}$  be a $1/ 4$-covering of $\Sphere{d_\theta-1}$ in the Euclidean norm with $|M|\leq 9^{d_\theta}$, and $\{\Direcb_1,\ldots,\Direcb_N\}$ to be a $1/ 4$-covering of $\Sphere{d_z-1}$ with $|N|\leq 9^{d_z}$. Then by a standard discretization argument, we have
\begin{align*}
    \opnorm{\frac{1}{n}\sum_{i=1}^n\theta_iz_{i}^\top-\E[\theta_i z_i]}\leq 2\sup_{k\in[M],l\in[N]}\frac{1}{n}\sum_{i=1}^n\Direc_{k}^\top \theta_iz_{i}^\top\Direcb_{l}-\E[\Direc_{k}^\top \theta_iz_{i}^\top\Direcb_{l}].
\end{align*}
Since $\Direc_{k}^\top \theta_iz_{i}^\top\Direcb_{l}-\E[\Direc_{k}^\top \theta_iz_{i}^\top\Direcb_{l}]$ are zero-mean subexponential variables by assumption, it follows that 
\begin{align*}
\P\left\{\left|\frac{1}{n}\sum_{i=1}^n\Direc_{k}^\top \theta_iz_{i}^\top\Direcb_{l}-\E[\Direc_{k}^\top \theta_iz_{i}^\top\Direcb_{l}]\right|\geq t\right\}\leq 2\exp\left(-c\min\left\{\frac{nt^2}{L_{\theta z}^2},\frac{nt}{L_{\theta z}}\right\}\right).
\end{align*}
Applying a union bound over $M,N$ and setting $t=cL_{\theta z}\sqrt{\frac{d_\theta+d_z+\log(1/\delta)}{n}}<cL_{\theta z}$ with some sufficiently large constant $c>0$ yields
\begin{align*}
\P\left\{\opnorm{\frac{1}{n}\sum_{i=1}^n\Direc_{k}^\top \theta_iz_{i}^\top\Direcb_{l}-\E[\Direc_{k}^\top \theta_iz_{i}^\top\Direcb_{l}]}\geq t\right\}
&\leq 
\P\left\{\sup_{k,l}\left|\frac{1}{n}\sum_{i=1}^n\Direc_{k}^\top \theta_iz_{i}^\top\Direcb_{l}-\E[\Direc_{k}^\top \theta_iz_{i}^\top\Direcb_{l}]\right|\geq t\right\}
\\
&
\leq 2\exp\left((d_\theta+d_z)\log 9-c\min\left\{\frac{nt^2}{L_{\theta z}^2},\frac{nt}{L_{\theta z}}\right\}\right)\leq \delta,
\end{align*}
which gives Eq.~\eqref{eq:linear_model_claim2}.

\section{Further experimental results and details}
\label{app:experiments}

We repeat each experiment $10$ times and plot the mean excess risk as well as the $\pm 1$ standard deviation. In all experiments on strategic classification, we choose the ridge parameter $\lambda=0.001$.

\begin{figure}[t]
  \centering
    \includegraphics[width=0.3\textwidth]{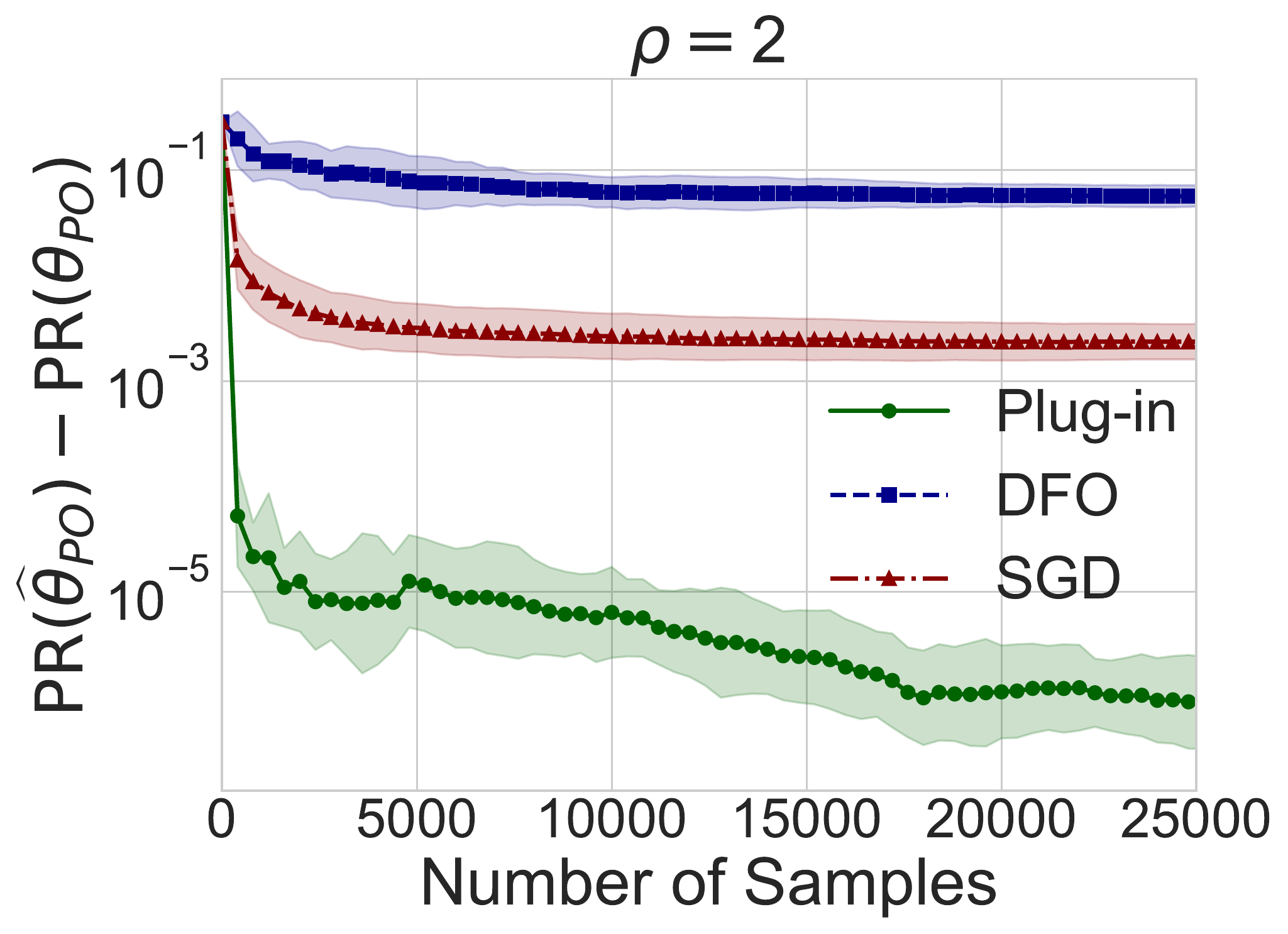}
      \includegraphics[width=0.3\textwidth]{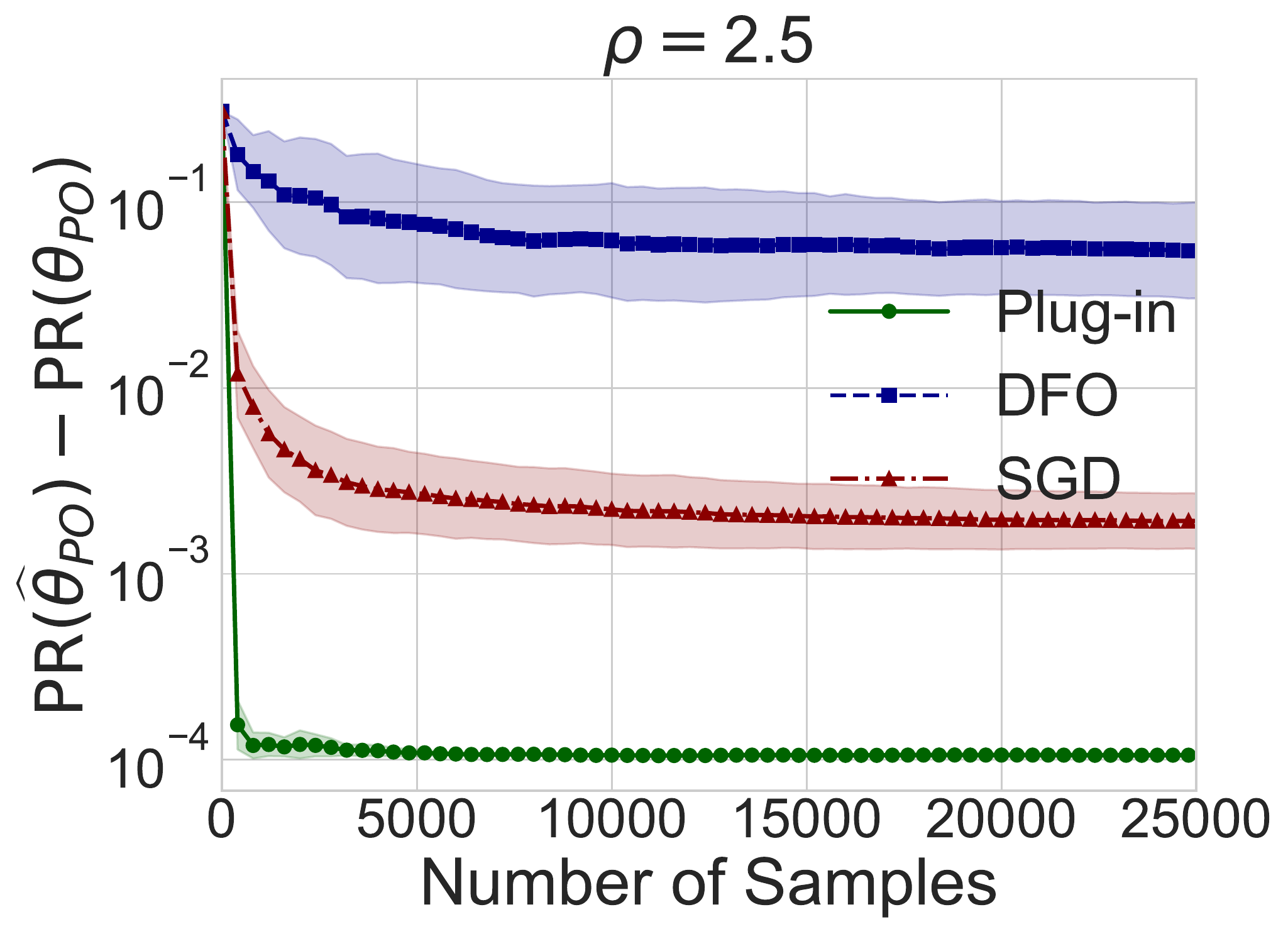}
  \includegraphics[width=0.3\textwidth]{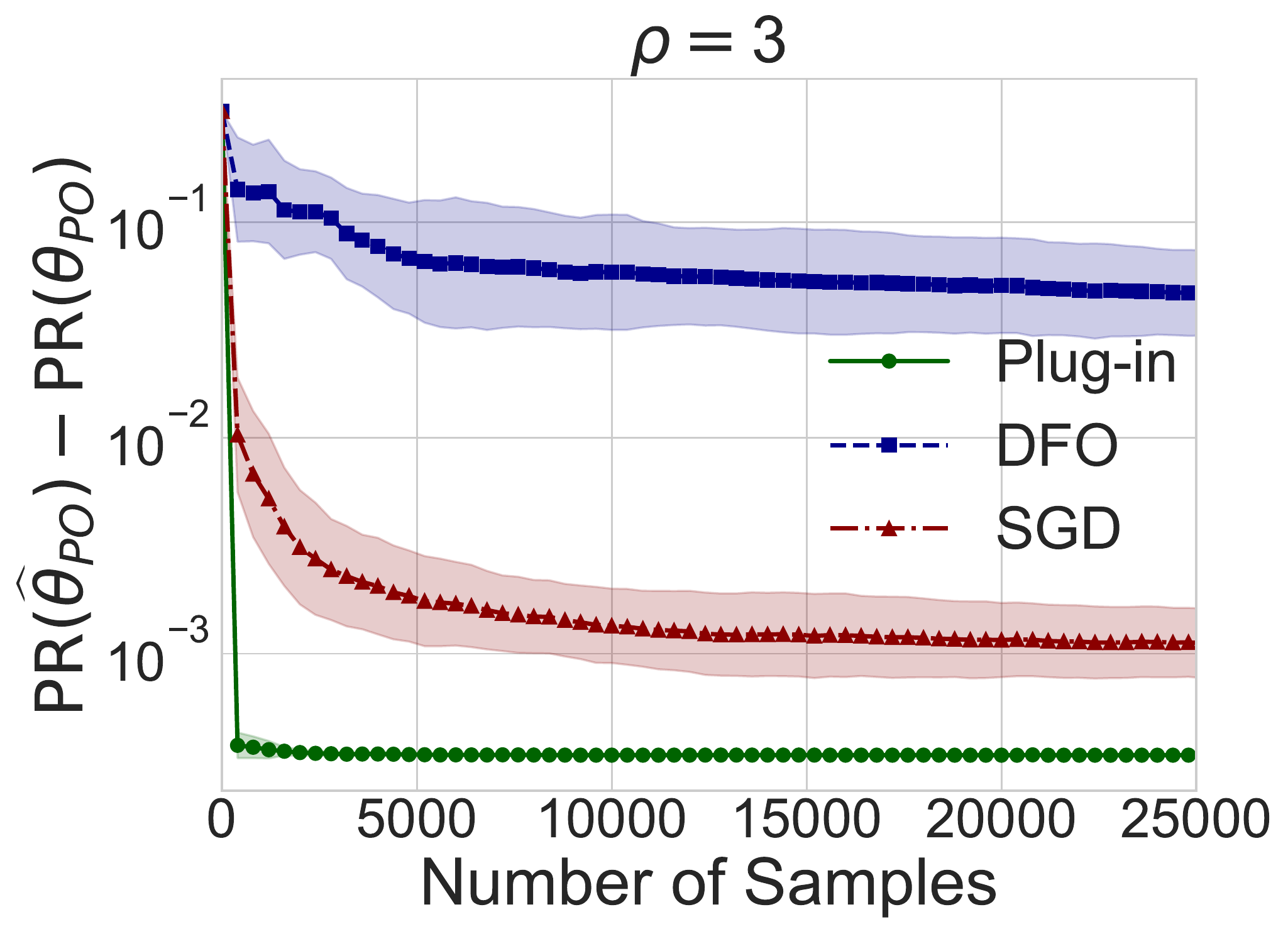}
   \includegraphics[width=0.3\textwidth]{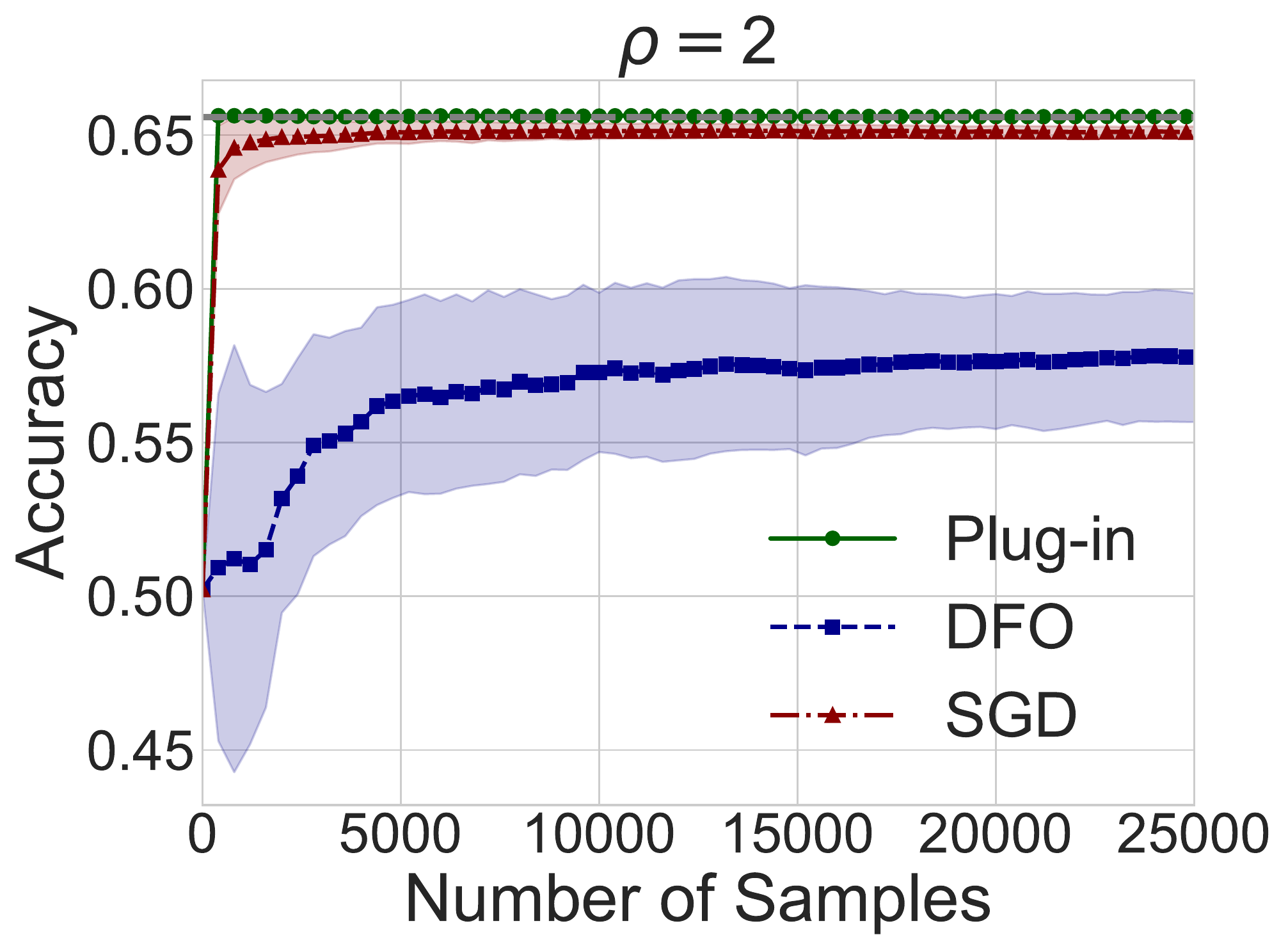}
     \includegraphics[width=0.3\textwidth]{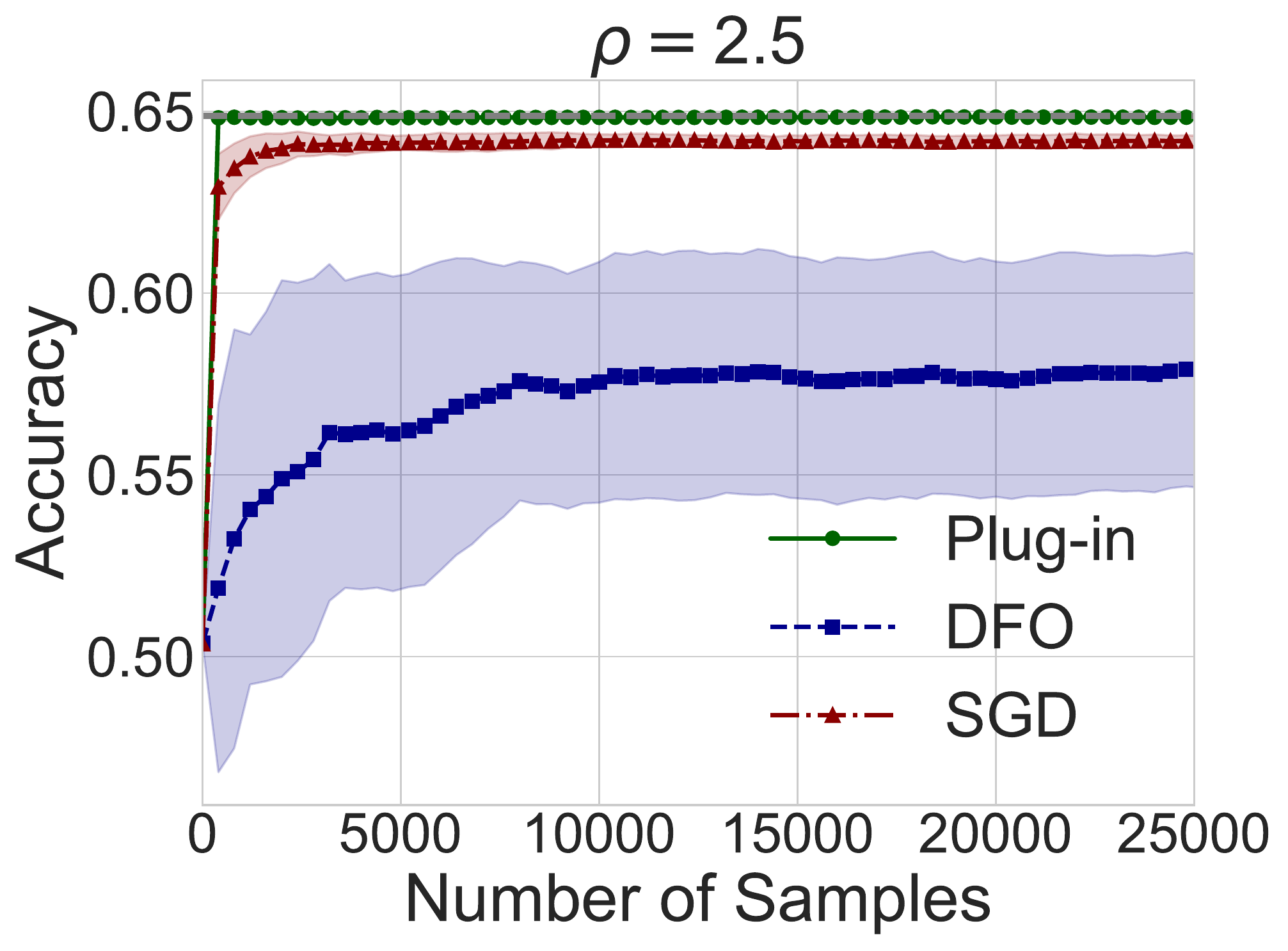}
  \includegraphics[width=0.3\textwidth]{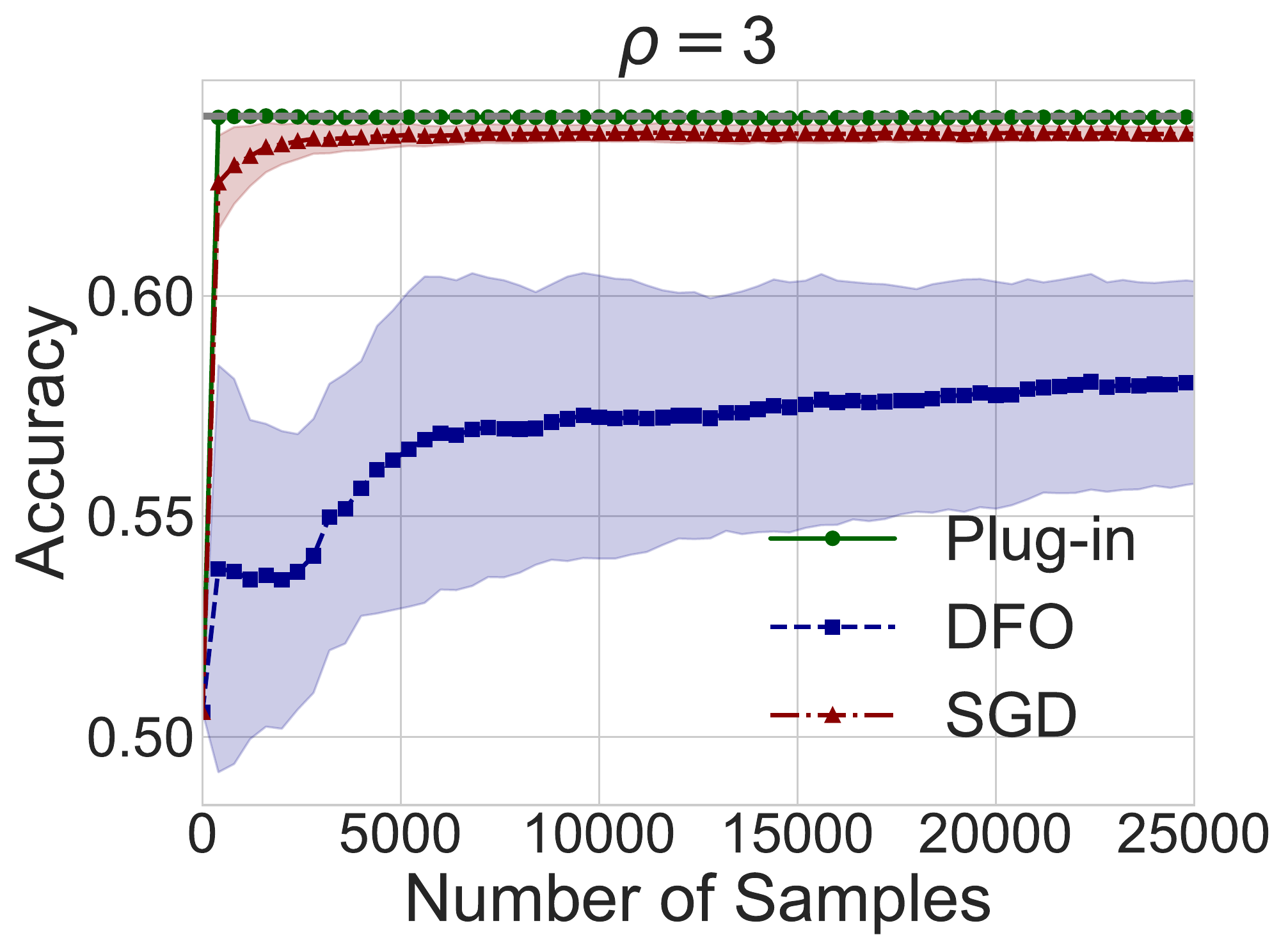}
  \caption{Excess risk (top) and accuracy (bottom) versus $n$ for plug-in performative optimization, the DFO algorithm, and greedy SGD, with a changed value of $\tilde\beta=1$.
 We display the $\pm 1$ standard deviation, logarithmically scaled. The takeaways are largely the same as in Figure \ref{fig:strat_regr_compare_1}.}
  \label{fig:strat_reg_additional}
\end{figure}

In Figure \ref{fig:strat_reg_additional} we provide an additional comparison in the context of the strategic-regression example from Section \ref{sec:experiments}. We let $\tilde\beta = 1$, showing that our takeaways are robust to the exact value of $\tilde\beta$.

\begin{figure}[t]
  \centering
    \includegraphics[width=0.3\textwidth]{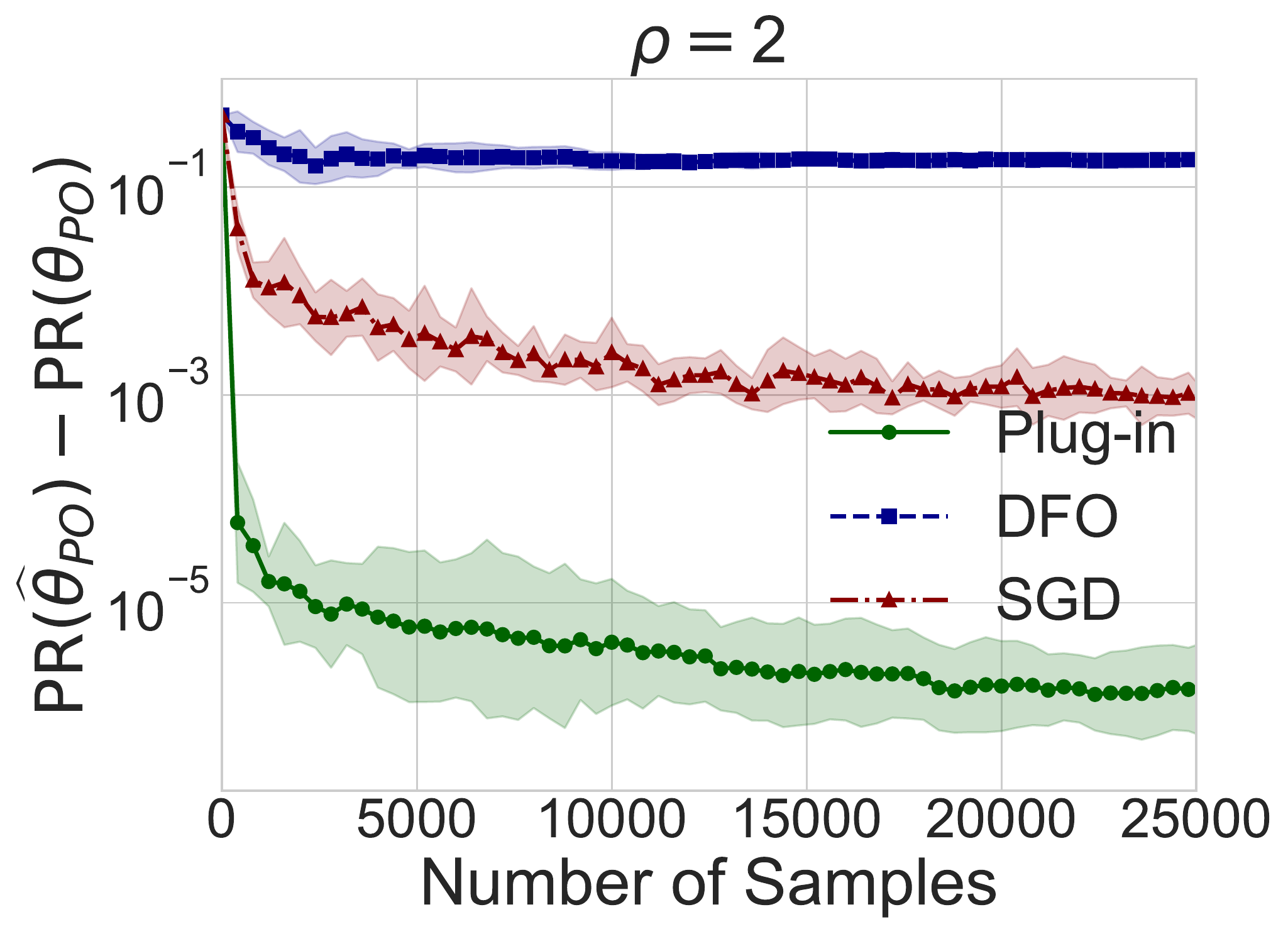}
     \includegraphics[width=0.3\textwidth]{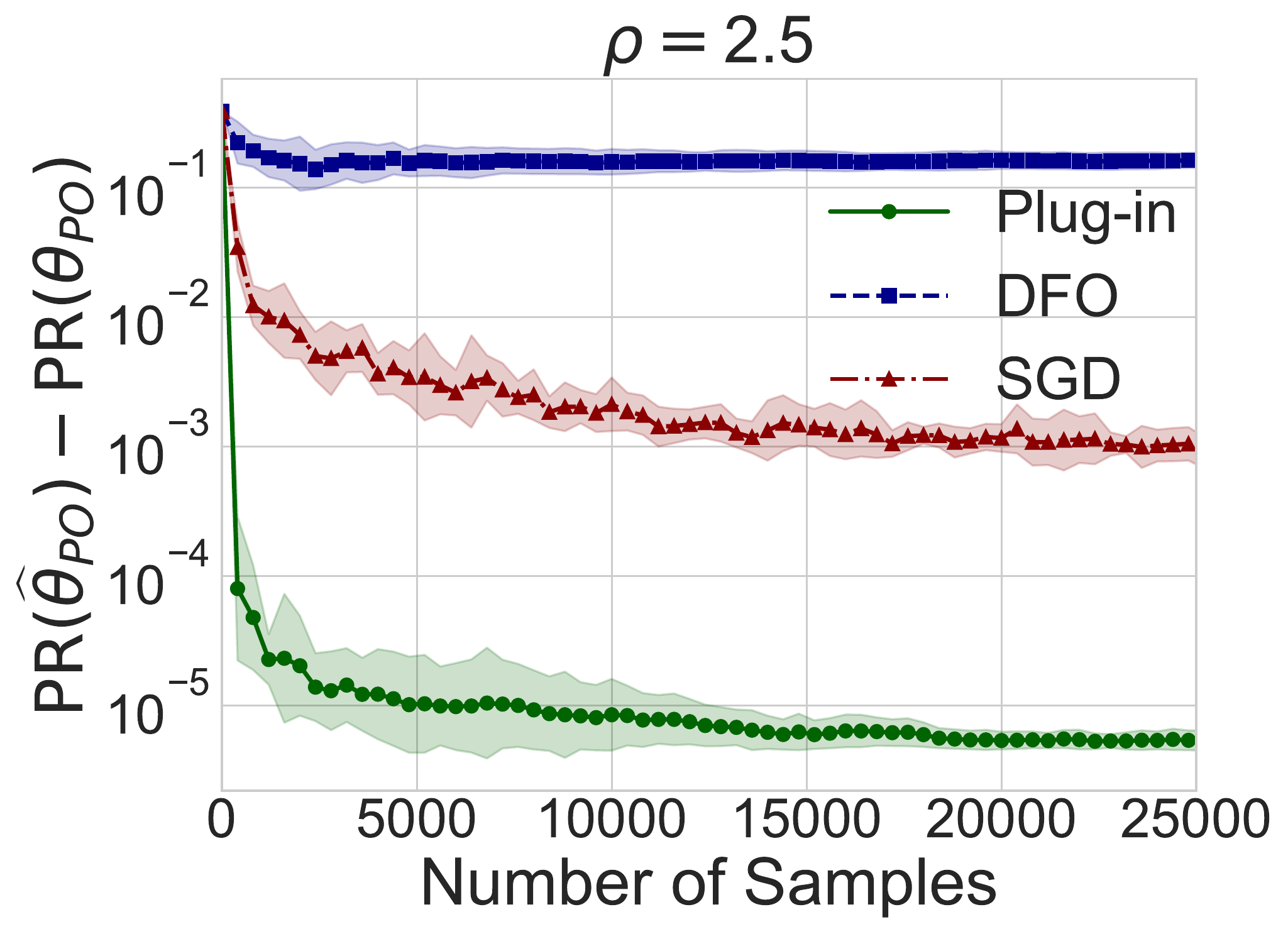}
        \includegraphics[width=0.3\textwidth]{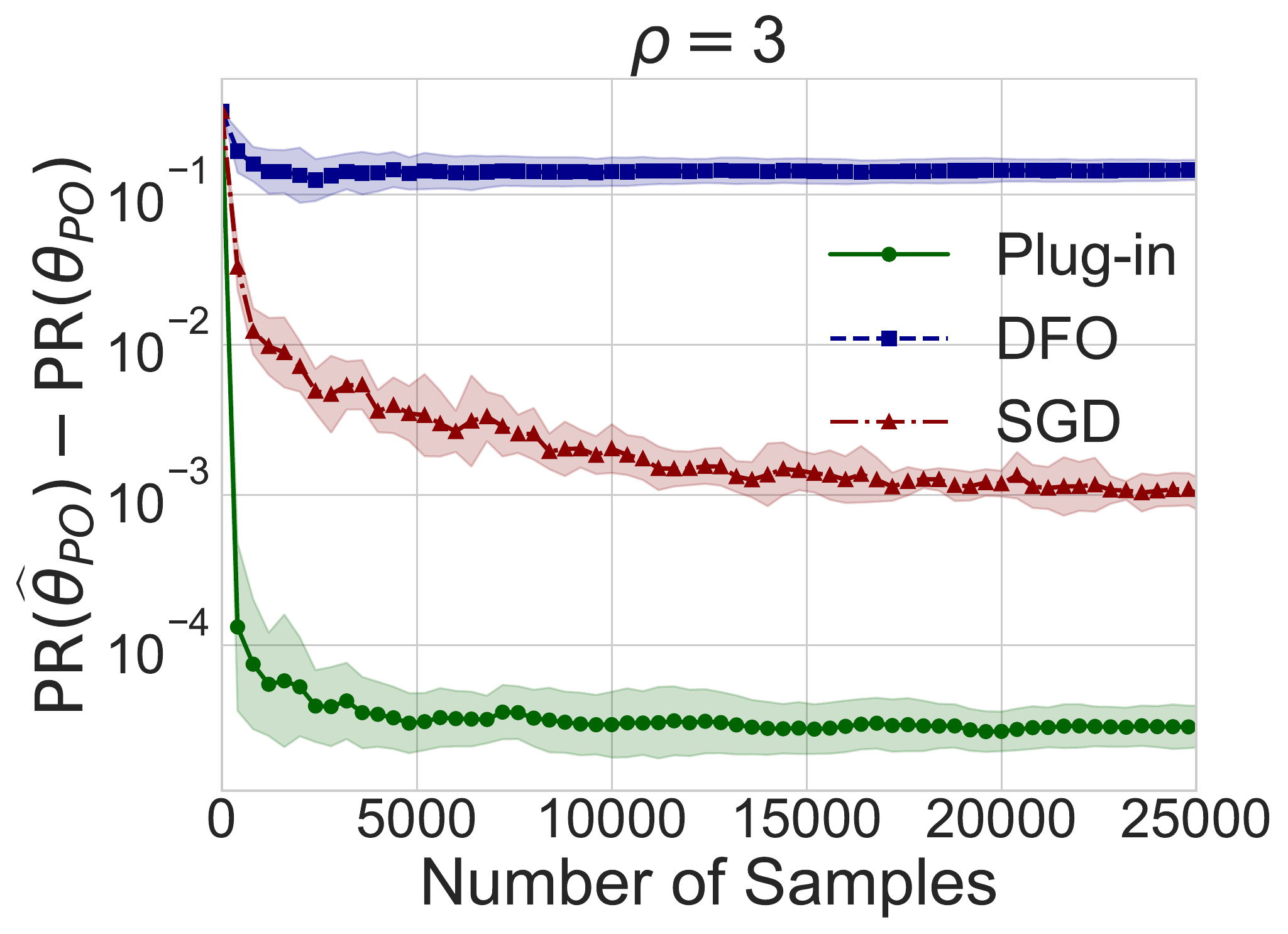}
        \includegraphics[width=0.3\textwidth]{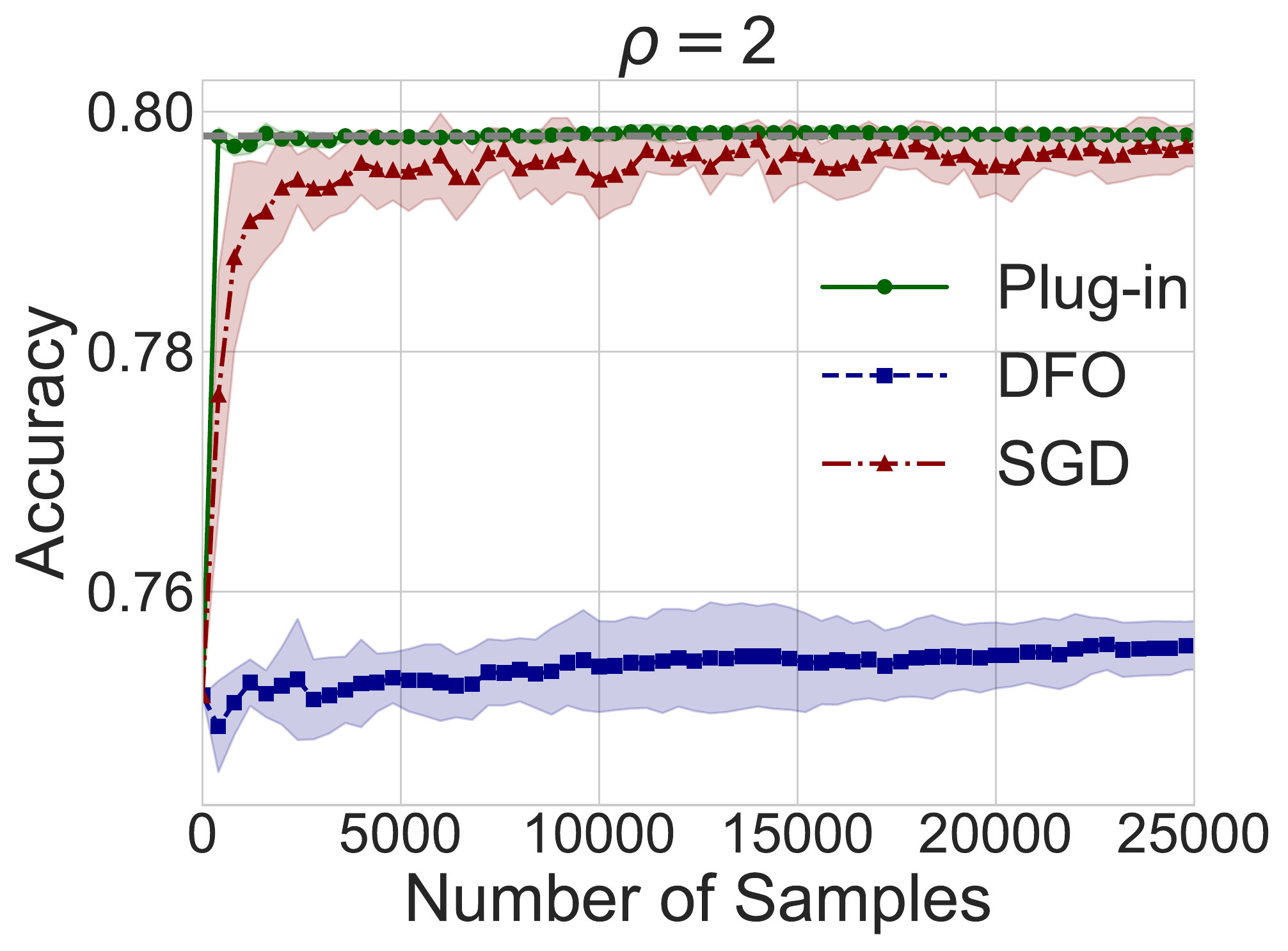}
         \includegraphics[width=0.3\textwidth]{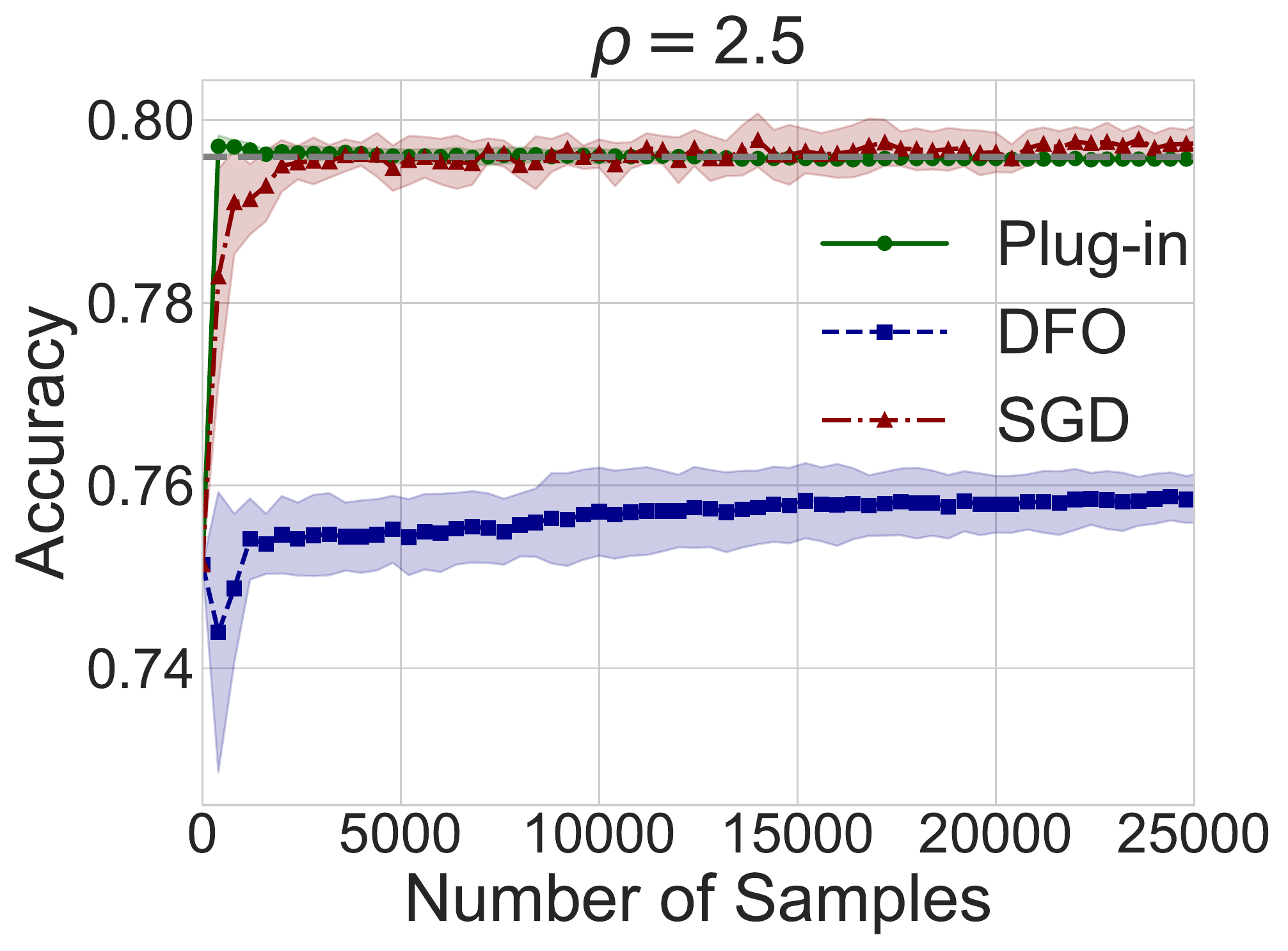}
 \includegraphics[width=0.3\textwidth]{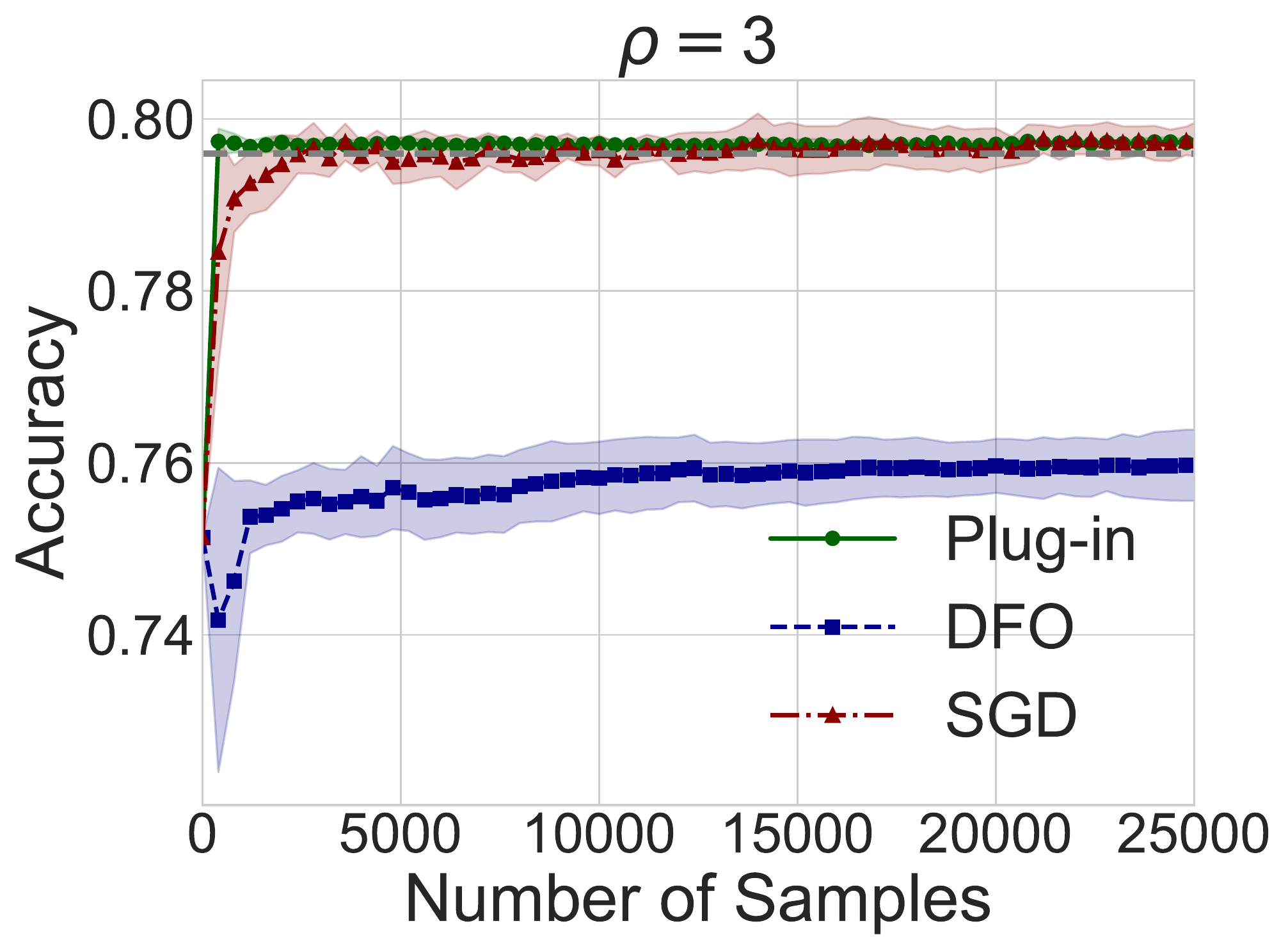}
  \caption{Excess risk (top) and accuracy (bottom) versus $n$ for plug-in performative optimization, the DFO algorithm, and greedy SGD, on the \texttt{credit} data set. 
 We display the $\pm 1$ standard deviation, logarithmically scaled.}
  \label{fig:real-data_additional}
\end{figure}

We also run the strategic-regression experiment on a real data set. We use the \texttt{credit} data set, in particular the processed version available at: \url{https://github.com/ustunb/actionable-recourse}. The data set contains $30,000$ samples of  $d=17$ features and a $\{0,1\}$-valued outcome $y_i$ with $y_i=1$ denoting individual $i$ not defaulting on a credit card payment. The features include marital status, age, education level, and payment patterns. We assume the individuals can modify their records on education level and payment patterns (features $7$--$17$), but cannot change other records. We use $1500$ randomly drawn data points to form the base distribution $\D_0$; we assume the same true response model and use the same distribution atlas as before.  We set $\tilde\beta=5$, $\Theta = \{\theta:\|\theta\|\leq 1\}$, and standardize the features so that each column is zero-mean and has unit variance. In  Figure~\ref{fig:real-data_additional}, we observe patterns similar to those in Figure~\ref{fig:strat_regr_compare_1}, though the gap in accuracy between our method and SGD is smaller.

Below we provide implementation details for the two considered baselines.

\paragraph{Derivative-free optimization (DFO).}  Starting from $\theta_0=\mathbf{1}_{d}/\sqrt{d}$, we run the updates
\[\theta_{t+1}=\mathrm{Proj}_{\enorm{\cdot}\leq 1}(\theta_t- \eta_t \widehat\E [\Direc\PR(\theta_t+\delta \Direc)d/\delta])\]
for $t\geq0$, where  the step size $\eta_t=c_0/(t+1)$, $\Direc$ is uniformly distributed on $\Sphere{d-1}$, and \mbox{$\widehat\E[\Direc\PR(\theta_t+\delta \Direc)d/\delta]$} denotes the unbiased sample estimation of $\E [\Direc\PR(\theta_t+\delta \Direc)d/\delta]$ using $m$ i.i.d. pairs of $(\Direc,z)\sim \mathrm{Unif}(\Sphere{d-1})\times \D(\theta_t+\delta\Direc)$.  The projection  $\mathrm{Proj}_{\enorm{\cdot}\leq 1}(x)$ denotes the projection of $x\in\R^{d}$ onto the ball $\{v\in\R^{d}:\enorm{v}\leq 1\}$ in Euclidean norm. We choose the step size parameter $c_0\in[10^{-4},10^{-1}]$, the  batch size $m$ in $[1,500]$, and $\delta\in[0.1,100]$ via grid search.

\paragraph{Greedy stochastic gradient descent (SGD).}  Starting from $\theta_0=\mathrm{1}_d/\sqrt{d}$, we run the updates 
\begin{align*}\theta_{t+1}=\mathrm{Proj}_{\enorm{\cdot}\leq 1}(\theta_{t}-\eta_t \nabla_{\theta}\ell(z_t;\theta_t))\end{align*} with step size
$\eta_t=c_0/(t+1)$ and $z_t\sim\D(\theta_t)$. The  step size parameter $c_0\in[10^{-4},10]$ and the batch size $m\in[1,500]$ are selected via grid search. The greedy SGD algorithm neglects the implicit dependence of $z$ on $\theta$ due to performativity, and therefore typically converges to suboptimal points.

\paragraph{Performative gradient descent (PerfGD).} Assume the distribution map has the form  $z\sim\D(\theta) \Leftrightarrow z\overset{d}{=} \cN(f(\theta),\sigma^2I_d)$, where $f$ is some unknown smooth function. Starting from $\theta_0=\mathrm{1}_d/\sqrt{d}$, we first run the greedy SGD updates for $\mathrm{H}$ burn-in steps. Next, we run SGD on the performative risk using an estimated performative gradient, namely,
\begin{align*}
\theta_{t+1}=\mathrm{Proj}_{\enorm{\cdot}\leq 1}(\theta_{t}-\eta_t \widehat{\nabla}_{\theta}\E[\ell(z_t;\theta_t)]),
\end{align*}
with step size
$\eta_t=c_0/(t+1)$ and $z_t\sim\D(\theta_t)$,
where the estimated performative gradient is computed as in Algorithm 3 and Eq.~(2) in~\cite{izzo2021learn} via numerically estimating the gradient $\frac{\partial f}{\partial\theta}$.

We choose the number of burn-in steps $\mathrm{H}=10d$.  The  step size parameter $c_0\in[10^{-4},10]$ and the batch size $m\in[1,500]$ are selected via grid search. PerfGD runs stochastic gradient descent on the performative risk using an estimated performative gradient. It should be noted that the numerical approximation of $\frac{\partial f}{\partial\theta}$ is unstable when $d>1$, which results in the suboptimal performance of PerfGD in our location-family experiment.

%\thetahat = \argmin_{\theta \in\Theta} \PR^{\hat\beta}(\theta) =  \argmin_{\theta \in\Theta} \E_{z\sim\D_{\hat\beta}(\theta)}[\ell(z;\theta)]
\section{Solving for \texorpdfstring{$\thetahat$}{theta}} 
\label{app:optimization}

The map $\D_{\hat\beta}$ belongs to the distribution atlas chosen by the learner, and as such, it is fully specified and known to the learner. Therefore, solving for $\thetahat  =  \argmin_{\theta \in\Theta} \E_{z\sim\D_{\hat\beta}(\theta)}[\ell(z;\theta)]$ can only incur error due to computational inaccuracies. There is no additional statistical complexity (i.e. dependence on $n$), which is the focus of our excess risk bounds in Theorem~\ref{thm:general_risk_bound}. In a sense, our results can be thought of as analogous to classical generalization bounds for empirical risk minimizers: we are concerned with characterizing the performance of the empirical risk minimizer, not with computational strategies for finding them.

More practically, there are several approaches one can take to compute $\thetahat$. Sometimes $\thetahat$ has a closed-form expression, as in Example~\ref{exm:biased_coin}. In such cases there is no error in Step 3 of Algorithm \ref{alg:general_procedure}.
Sometimes $\D_{\hat\beta}(\theta)$ and $\ell(z;\theta)$ are simple enough that $\mathrm{PR}^{\hat\beta}(\theta)$ has a closed-form expression; in such cases, we compute $\thetahat$ by running gradient descent on $\PR^{\hat\beta}(\theta)$.  This is the case in all our experiments.  Alternatively, if $\PR^{\hat\beta}(\theta)$ does not have a closed-form expression, one may compute $\thetahat$ by using a black-box optimizer on an unbiased estimate of $\E_{z\sim\D_{\hat\beta}(\theta)}[\ell(z;\theta)]$ obtained by drawing many i.i.d. samples from $\D_{\hat\beta}(\theta)$ (the right algorithm depends on what we know about the problem; generically we can always use DFO \cite{Flaxman2004OnlineCO}). Since these samples are all synthetic and \emph{do not} count toward the sample complexity---i.e., they do not require collecting real data but only simulation---we can draw arbitrarily many samples to achieve a small numerical error.

\end{document}